\documentclass{article}

\usepackage[preprint]{neurips_2026}

\usepackage[utf8]{inputenc}   
\usepackage[T1]{fontenc}      
\usepackage{url}              
\usepackage{booktabs}         
\usepackage{amsfonts}         
\usepackage{nicefrac}         
\usepackage{microtype}        
\usepackage{graphicx}         

\usepackage{float}            
\usepackage{enumitem}
\usepackage{needspace}
\usepackage{multirow}
\usepackage[table]{xcolor}
\usepackage{makecell}
\usepackage{tabularx}
\usepackage{pifont}
\usepackage{amsmath}
\usepackage[hidelinks]{hyperref} 

\definecolor{ourblue}{HTML}{ECF4FF}

\title{CleanMDM: Clean Motion Diffusion Model for Multimodal Motion Cleanup}

\author{%
  Zhe Li$^{1,*}$ \quad
  Shicheng Wang$^{2,*}$ \quad
  Bowen Cai$^{3}$ \quad
  Huan Fu$^{3,\dagger}$ \\
  $^{1}$Peking University \quad
  $^{2}$Light Origins \quad
  $^{3}$Alibaba \\
  $^{*}$Equal contribution. \quad
  $^{\dagger}$Corresponding author.
}

\hypersetup{
  pdftitle={CleanMDM: Clean Motion Diffusion Model for Multimodal Motion Cleanup},
  pdfauthor={Zhe Li, Shicheng Wang, Bowen Cai, Huan Fu}
}

\begin{document}

\maketitle

\vspace{-4mm}
\begin{figure}[H]
  \vspace{-8pt}
  \centering
  \includegraphics[width=\textwidth]{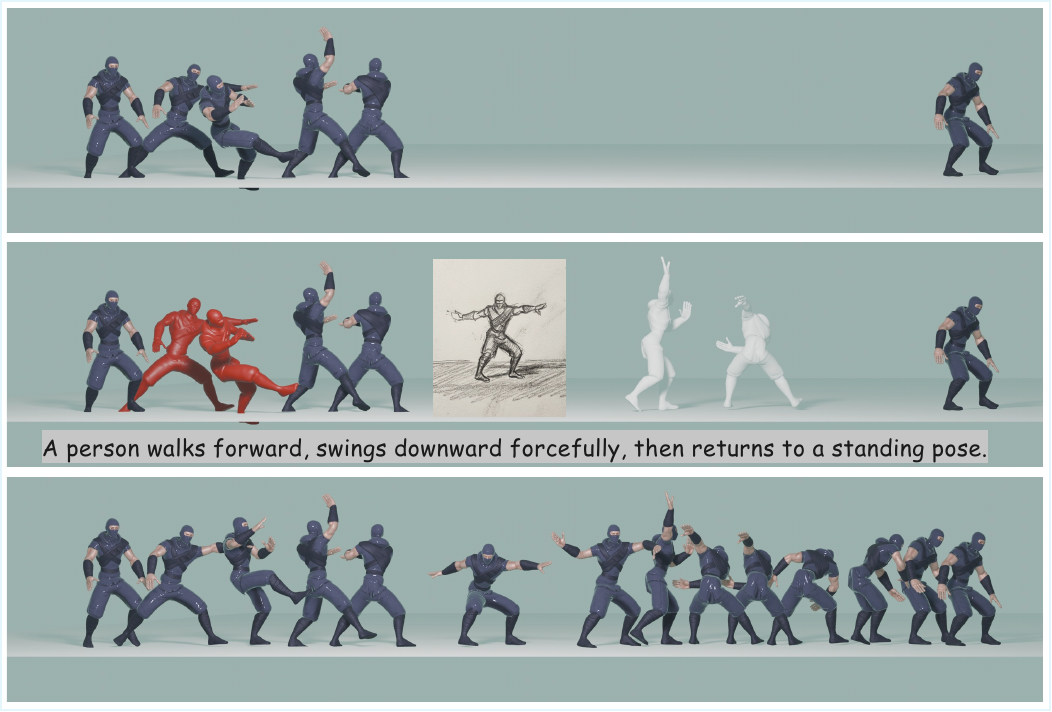}
  \vspace{-14pt}
  \vspace{-2mm}
  \caption{\textbf{Overview of motion cleanup.}
    \textbf{Top:} the input motion with a localized corrupted interval, represented as masked noisy motion.
    \textbf{Middle:} the same clip augmented with optional, plug-and-play conditions: corrupted frames are highlighted in red for visualization, a reference image provides sparse 2D keyframes, white silhouettes indicate sparse 3D keyframes, and a text prompt specifies high-level intent.
    \textbf{Bottom:} CleanMDM generates a clean and temporally consistent motion sequence.}
  \label{fig:teaser}
  \vspace{-2mm}
\end{figure}

\begin{abstract}
Motion capture data is rarely directly usable, as they typically exhibit missing segments, jitter, drift and contact artifacts.
Traditionally, corrupted motions are cleaned by animators through the manual identification of keyframes from noisy motion, subsequent keyframe correction, and interpolation between corrected keyframes to reconstruct coherent motion.
While the rise of generative motion models has made automatic cleanup feasible, most approaches operate as black box denoisers with limited controllability, making it difficult to preserve reliable segments or enforce specific user intents.
Inspired by animation workflows, we present \textbf{CleanMDM}, a unified multimodal motion cleanup framework that formulates cleanup as masked conditional generation with plug-and-play conditions.
This single model supports arbitrary combinations of noisy 3D motion, sparse 2D keyframes, sparse 3D keyframes, and text.
This design enables both automatic cleanup without additional user annotation and controllable cleanup under multimodal guidance.
To further improve motion realism, we incorporate the Latent Motion Quality Discriminator (\textbf{LMQD}) to better match kinematic distributions and reduce skating, jitter, and interpenetration artifacts, and we apply \textbf{Mesh-Aware Contact Projection} as a test-time optimization step to enhance contact and physical consistency.
Experiments across multiple datasets demonstrate that \textbf{CleanMDM} consistently outperforms prior cleanup and generation baselines, and that low cost conditions (text and 2D keyframes) provide reliable controllability gains in multimodal cleanup scenarios.
\end{abstract}

\vspace{-1mm}
\section{Introduction}
\label{sec:intro}


Motion data in modern animation pipelines is increasingly acquired from video-based motion capture~\cite{shen2024world,goel2023humans} and generative motion models~\cite{tevet2023human,chen2023executing}, which are more accessible and scalable than traditional studio capture but frequently introduce missing observations, temporal jitter, drift, and contact artifacts such as foot sliding, floating, and ground penetration.
Because these artifacts are visually amplified after skinning and rendering, animators still spend substantial time on manual cleanup and inbetweening, making motion repair a persistent production bottleneck~\cite{williams2012animator,holden2018robust,kovar2002footskate}.


A key characteristic of real-world motion corruption, largely overlooked by existing methods, is that degradation is almost always localized: only certain temporal intervals or body parts are affected, while the remaining motion stays reliable.
For example, video-based estimates may exhibit lower-body drift while the upper body is essentially correct, or jitter may appear in only a handful of joints over a short window.
This partial reliability has a direct consequence for method design: an ideal cleanup system should identify and preserve the trustworthy portions of a motion while selectively regenerating only the corrupted regions, and it must accept diverse forms of guidance depending on what information is available for a given shot.


Existing approaches largely fall into two categories, neither of which is specifically designed to handle partial reliability.
\textbf{Restoration-style denoisers}~\cite{mu2025stablemotion,zhang2024rohm} map an entire noisy sequence to a clean motion in a black-box fashion.
They improve average reconstruction error but treat every frame uniformly, providing limited controllability when animators need to lock key poses, preserve trustworthy segments, or repair only a specified interval.
\textbf{Inbetweening and conditional generation}~\cite{cohan2024flexible,zhang2025plug,meng2025absolute} take the opposite stance: they regard the corrupted region as entirely unobserved and reconstruct it from sparse constraints (keyframes, text).
When applied to cleanup, this discards the partially reliable observations inside the corrupted interval, leading to over-reconstruction, style drift, and unstable contact behavior.
Moreover, these two families of methods are typically developed and deployed as separate, modality-specific models, which fragments supervision, limits cross-modal knowledge transfer, and requires maintaining multiple pipelines.
It is worth noting that motion cleanup is fundamentally different from motion generation: generation creates motion from scratch given sparse or abstract signals, whereas cleanup starts from an existing clip that is partially reliable and must preserve its trustworthy content.
Recent multimodal generation models such as ~\cite{li2025genmo} accept heterogeneous inputs (2D/3D keyframes, text) but are designed for the generation task. They do not model per-frame reliability or enforce non-destructive preservation of the input draft, which limits their applicability to production cleanup scenarios requiring fidelity to the original performance.


We present \textbf{CleanMDM}, a unified multimodal motion cleanup framework that directly addresses the partial-reliability challenge.
Our key insight is to formulate cleanup as \emph{masked conditional generation} in a shared latent space.
The central mechanism is \emph{confidence-aligned multimodal fusion}.
Heterogeneous conditions, including noisy 3D motion drafts with per-frame confidence masks, sparse 2D keyframes, sparse 3D keyframes, and text prompts, are encoded into a common latent representation and fused via frame-wise confidence weighting (Eq.~\ref{eq:weighted_fusion}).
Unlike raw-space additive fusion used in generation models~\cite{li2025genmo}, our confidence-aligned design explicitly encodes the \emph{reliability} of each frame: high-confidence frames are preserved, low-confidence frames are regenerated, and the boundary between the two is determined by the per-frame weights rather than a hard binary mask.
This enables a smooth continuum from fully missing intervals (inbetweening) to partially degraded segments (denoising-style cleanup), all within a \emph{single} model.
Text and sparse 2D keyframes serve as low-cost, plug-and-play conditions that provide semantic and geometric guidance without requiring expensive 3D annotation.
By jointly training on all modalities, CleanMDM encourages cross-modal transfer: the richer supervision yields a stronger motion prior that benefits even the simplest auto-cleanup setting where no additional guidance is provided.


Two additional technical contributions improve motion quality beyond standard regression objectives.
First, we introduce the \textbf{Latent Motion Quality Discriminator (LMQD)}, a lightweight adversarial regularizer that operates on entire sequences in the motion autoencoder's latent space.
Unlike prior kinematic-feature discriminators used in physics-based control~\cite{peng2021amp,luo2023perpetual}, LMQD back-propagates gradients directly through the latent representation, better capturing distributional characteristics of natural motion and more effectively penalizing perceptually salient artifacts (skating, jitter, penetration) that have small $\ell_2$ magnitude but high visual salience.
Second, we propose \textbf{Mesh-Aware Contact Projection}, a test-time optimization step that refines the full sequence with explicit constraints on SMPL mesh foot-sole vertices.
Unlike conventional joint-level inverse kinematics~\cite{holden2018robust,yao2024moconvq}, our formulation reasons about body-surface geometry relative to the ground plane, enforcing hard contact consistency that joint-only approaches structurally cannot achieve.




Our contributions are summarized as follows:
\begin{itemize}[leftmargin=1.5em, itemsep=2pt, topsep=2pt]
    \item \textbf{Confidence-aligned multimodal cleanup framework.}
    We formulate motion cleanup as masked conditional generation in a shared latent space with per-frame confidence weighting, enabling non-destructive repair that preserves reliable motion while regenerating corrupted regions --- a capability that neither black-box denoisers nor generation-oriented models provide. A single CleanMDM model covers auto cleanup, inbetweening, and multimodal-guided repair.
    \item \textbf{Latent adversarial regularization.}
    We introduce LMQD, a latent-space sequence discriminator that provides distributional supervision beyond regression losses, yielding measurable improvements on contact and temporal-quality metrics.
    \item \textbf{Mesh-aware contact projection.}
    We propose a test-time optimization that constrains SMPL mesh vertices to enforce surface-level ground contact (foot sliding, penetration, and floating).
\end{itemize}

\vspace{-1mm}
\section{Related Works}
\label{sec:relatedworks}
\vspace{-3mm}
\subsection{Human Motion Generation}
\vspace{-3mm}
Recent progress in human motion synthesis is largely driven by generative models. Diffusion approaches (e.g. \cite{tevet2023human,chen2023executing,zhang2024motiondiffuse,zhao2024dartcontrol,zhang2024motion,zhang2023remodiffuse}) show their ability to model long-range spatio-temporal motion relationships, while VQ/tokenized pipelines \cite{zhang2023generating,guo2024momask,li2024lamp,jiang2023motiongpt,pinyoanuntapong2024mmm,zhang2024tedi} introduce discrete motion token modeling.
Representations evolve from kinematics-rich descriptors (e.g., HumanML3D \cite{guo2022generating}) toward more concise forms, including redundancy reduced designs \cite{meng2024rethinking} and absolute coordinate modeling \cite{meng2025absolute}. Control has expanded beyond text to geometric cues and interactive editing interfaces \cite{wang2025stickmotion,zhong2025sketch2anim,karunratanakul2023guided,xie2023omnicontrol}, and more recently to multimodal guidance \cite{li2025genmo,luo2024m,xu2025omnimotion,zhang2025motion,chen2025motion}; some works also pair generation with physics for executability \cite{yuan2023physdiff,tevet2024closd,tessler2024maskedmimic,fan2025go,wang2025skillmimic}. 
Despite these advances, semantic conditions often underconstrain fine geometry and contacts, while physics constraints can disrupt motion timing and cause local limb deformations.


\vspace{-3mm}
\subsection{Motion Cleanup}
\vspace{-3mm}

Repairing motion captured from videos or produced as drafts is critical for production.
Existing work largely follows two directions. 
Cleanup-style methods map corrupted sequences to clean motion via end-to-end denoising \cite{mu2025stablemotion,zhang2024rohm}, reducing reconstruction error but offering limited fine-grained control (e.g., preserving reliable segments or enforcing locked poses). 
Some approaches detect corrupted segments within a motion sequence and subsequently regenerate them. However, these approaches treat missing regions as uninformative and do not explicitly combine partially reliable observations with heterogeneous constraints. In this sense, inbetweening can be viewed as a repair mechanism, as inbetweening and conditional completion synthesize motion from sparse constraints such as keyframes or text \cite{cohan2024flexible, zhang2025plug}.
In real world cleanup, corruption is often localized, and ignoring valid observations can increase ambiguity, causing over-reconstruction, style drift, and unstable contact-related artifacts, which are central to industrial requirements.
Recently, GenMo~\cite{li2025genmo} unifies generation and estimation with multimodal inputs, but as a generation model it does not model per-frame reliability.

\vspace{-3mm}
\subsection{Motion Quality and Physical Plausibility}
\vspace{-3mm}

To mitigate motion artifacts such as sliding, floating, and penetration, prior work incorporates contact losses~\cite{tevet2023human,tevet2024closd}, physics-driven refinement~\cite{luo2023perpetual,yao2024moconvq}, IK post-processing~\cite{shen2024world}, and adversarial motion priors (e.g., AMP~\cite{peng2021amp}).
These approaches address complementary aspects of motion quality; achieving both controllable repair and physical consistency typically requires combining learned priors with explicit contact enforcement, ideally avoiding heavy simulation during training.



\begin{figure*}[t]
  \centering
  \includegraphics[width=1\textwidth]{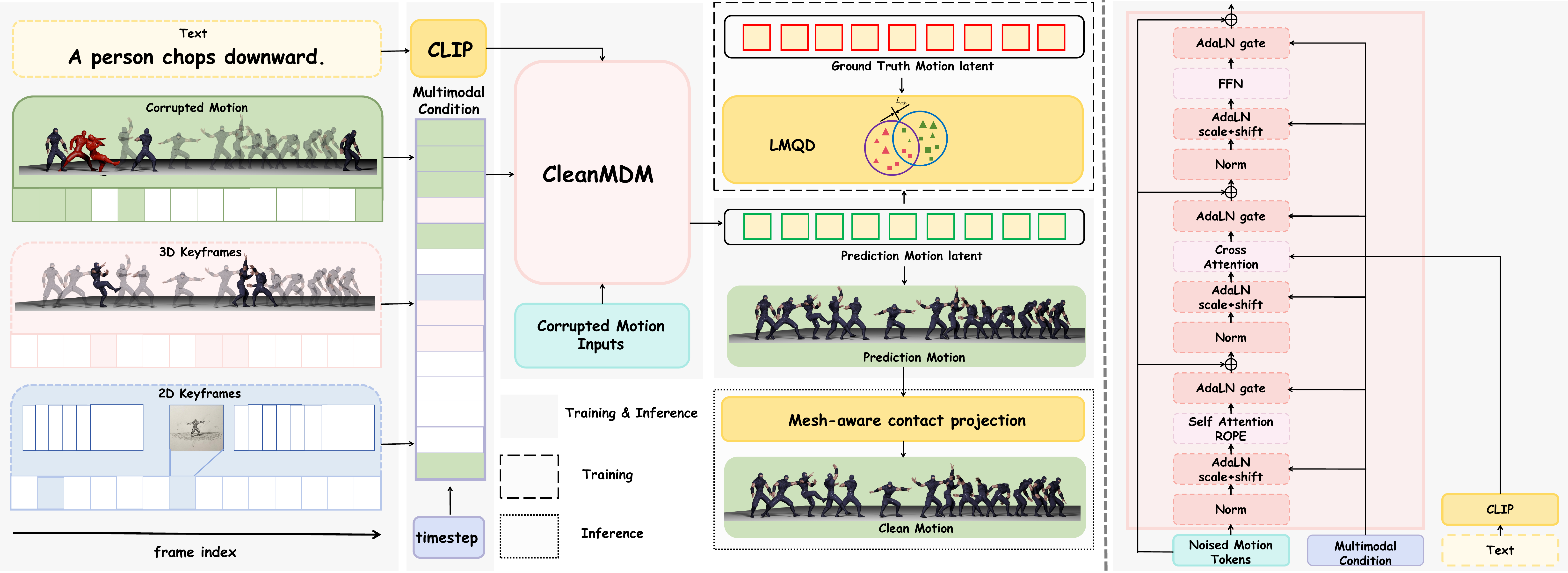}
    \vspace{-4mm}
    \caption{
    \textbf{CleanMDM framework and architecture.}
    \textbf{Left:} CleanMDM formulates motion cleanup as masked conditional generation in the autoencoder latent space, using a noisy 3D draft with frame-wise confidence and optional plug-and-play cues (sparse 2D/3D keyframes and text).
    Solid-bordered inputs are required; dashed-bordered inputs are optional.
    \textbf{Visualization:} semi-transparent skeletons indicate masked-out (unobserved) regions, and red skeletons indicate corrupted frames in the noisy draft.
    LMQD is used only during training for latent adversarial regularization, and an optional Mesh-Aware Contact Projection post-process is applied at inference to reduce contact artifacts.
    \textbf{Right:} CleanMDM block with frame weighted geometric fusion injected via AdaLN+gated residuals into self-attention (RoPE) and FFN; text is injected through cross-attention.
    \vspace{-4mm}
    }

  \label{fig: fig_framework}
\end{figure*}

\vspace{-1mm}
\section{Data Representation}
\label{sec:data}

\vspace{-3mm}
We represent a motion clip as absolute joint positions
$\mathbf{X}_{1:T}\in\mathbb{R}^{T\times J\times 3}$ with $J{=}22$.
Following ACMDM~\cite{meng2025absolute}, we adopt absolute coordinates as the interface for cleanup because they align naturally with the spatial signals used in our task.
Sparse 3D keyframes can be inserted in the same coordinate system, 2D keypoints are obtained by direct camera projection, and contact reasoning or inference-time IK can be formulated directly in metric space.
Compared with relative or velocity-based representations, absolute coordinates also avoid reconstructing global trajectories by accumulating local offsets, where small frame-wise errors can propagate into long-horizon drift.
This makes the representation especially suitable for non-destructive cleanup: reliable observations, sparse controls, and regenerated regions all remain comparable in a single world-space coordinate system.

All conditions are aligned to the latent temporal resolution produced by the motion autoencoder~\cite{kingma2013auto}.
With a downsampling factor of $s{=}4$, the latent length is $T'=\lceil T/s\rceil$.

\vspace{-3mm}
\paragraph{Noisy 3D observations.}
We denote the corrupted joint sequence by $\tilde{\mathbf{X}}_{1:T}$.
A frame-level corruption indicator or user-provided interval annotation is converted into a confidence weight and applied to $\tilde{\mathbf{X}}_{1:T}$ before encoding, so unreliable frames contribute less to the noisy 3D condition.
After encoding, we obtain noisy 3D condition tokens and their latent-level weights $w_{\text{noisy}}\in[0,1]^{T'}$.

\vspace{-3mm}
\paragraph{Sparse clean 3D keyframes.}
From frames marked as unreliable, we randomly sample a portion as clean 3D keyframes during training.
These frames provide sparse ground-truth constraints inside degraded intervals.
We store a binary keyframe mask and its latent-level weights $w_{\text{clean}}\in[0,1]^{T'}$.

\vspace{-3mm}
\paragraph{2D keypoints.}
We use 2D keypoints as a dense geometric condition rather than a supervised loss.
For each clip, a camera path is sampled from \{\textit{fixed}, \textit{dolly in}, \textit{dolly out}, \textit{follow}, \textit{orbit}\} to obtain extrinsics $(\mathbf{R}_t,\mathbf{p}_t)$; fixed intrinsics $K$ are used for perspective projection.
The 2D sequence is fed as the conditioning signal $\mathbf{U}_{1:T}$.
We apply a standard crop and scale normalization to absolute image scale.
During training, we mimic sparse detections by sampling a small portion of unreliable frames as 2D keyframes.
Their latent-level weights are denoted by $w_{\text{2D}}\in[0,1]^{T'}$.
\vspace{-1mm}
\begin{equation}
    \mathbf{u}_{t,j}=\Pi(K,\mathbf{R}_t,\mathbf{p}_t;\mathbf{X}_{t,j}),\ t\in[1{:}T],\ j\in[1{:}J].
\end{equation}

\vspace{-5mm}
\paragraph{Text.}
Each clip is paired with a textual description.
We tokenize the caption and extract token-level features using a frozen CLIP text encoder~\cite{radford2021learning}.
These features are used as an additional conditioning context via cross-attention.

\vspace{-1mm}
\section{Method}
\label{sec:method}

\vspace{-1mm}
CleanMDM formulates motion cleanup as masked conditional generation in a shared latent space.
The central design principle is confidence-aligned multimodal fusion: heterogeneous conditions are encoded into a common latent representation and fused via per-frame confidence weighting, so that trustworthy frames are preserved while corrupted regions are selectively regenerated.
This fundamentally distinguishes CleanMDM from generation models (e.g., GenMo~\cite{li2025genmo}) that synthesize motion from scratch without modeling input reliability.

\vspace{-1mm}
\subsection{Latent Motion Modeling Backbone}
\vspace{-1mm}

CleanMDM performs motion cleanup in a learned latent space.
Given an absolute joint trajectory $\mathbf{X}_{1:T}\in\mathbb{R}^{T\times J\times 3}$, we encode it with a motion autoencoder $\mathcal{E}$ and decode with $\mathcal{D}$.
The encoder downsamples only in time by a factor $s{=}4$, producing a latent sequence of length $T'=\lceil T/s\rceil$ while keeping the joint dimension $J$ unchanged.
We train a conditional flow matching Transformer \cite{lipman2022flow, peebles2023scalable, ho2020denoising} to predict a velocity field in latent space, and we sample with an ODE solver \cite{chen2018neural}. Working in latent space reduces computation, improves training stability, and allows us to integrate heterogeneous conditions at a common temporal resolution.
Our backbone is a conditional Transformer over the noised latent motion tokens.
We use RoPE positional encoding \cite{su2024roformer} and QK Normalization \cite{henry2020query} in self attention, and inject conditions by AdaLN modulation with residual gating, which allows the model to smoothly activate or deactivate guidance under missing or partial conditions.
Text features from a frozen CLIP encoder are provided as context to a gated cross attention layer.

\vspace{-1mm}
\subsection{Multimodal Conditions and Fusion}
\vspace{-1mm}
CleanMDM treats motion cleanup as masked conditional generation with plug-and-play conditions.
The core mechanism is \emph{confidence-aligned fusion in latent space}: instead of directly concatenating raw inputs or using additive fusion in joint space, we encode all geometric conditions through the shared autoencoder and fuse them via per-frame confidence weighting.
This design is motivated by two reasons.
First, the autoencoder provides a shared coordinate system and scale for 3D motion signals, which makes fusion well-posed and improves robustness to partial corruptions.
Second, latent fusion follows the successful recipe of latent diffusion in vision, where rich conditions are projected to the model width and injected into a generative backbone without forcing all modalities into a brittle raw-space representation.
Unlike raw-space additive fusion used in generation models~\cite{li2025genmo}, our confidence-aligned design explicitly encodes the reliability of each frame, enabling non-destructive cleanup.

\vspace{-3mm}
\paragraph{Condition construction}
For 3D conditions, we \emph{mask in input space first}, then encode with the same $\mathcal{E}$:
(i) noisy 3D observations use a frame-wise confidence derived from an interval annotation or automatic corruption indicator so unreliable intervals contribute less;
(ii) sparse clean 3D keyframes use a binary mask so only selected frames are kept.
The resulting latents are flattened into per-time tokens, and their masks are downsampled to the latent rate to obtain frame weights.
For 2D keyframes, we mask in 2D space, downsample to $T'$, flatten each frame, and map it with a small MLP, without explicit 2D-to-3D lifting.
Since the backbone learns a 3D motion latent from paired projected 2D cues, the MLP only needs to preserve image-plane geometry while the motion prior infers compatible 3D motion.
This avoids the feature-space mismatch introduced by heavier pretrained 2D representations; Appendix~\ref{sec:ablation_condition_encoders} (Table~\ref{tab:ablation_2d_encoder}) confirms that it outperforms MotionBERT features.

\vspace{-3mm}
\paragraph{Weighted fusion}
Cleanup differs from generation because the input draft is partially reliable.
StableMotion~\cite{mu2025stablemotion} targets corrupted motion but lacks a unified multimodal interface, while GenMo~\cite{li2025genmo} accepts heterogeneous inputs but does not encode which draft frames are trustworthy.
We therefore expose reliability as frame weights, preserving high-confidence draft frames while letting optional keyframes dominate unreliable intervals.

Let $\mathbf{y}^{\text{noisy}}_t$, $\mathbf{y}^{\text{clean}}_t$, and $\mathbf{y}^{\text{2D}}_t$ be the condition tokens at latent frame $t$,
and let $w_{\text{noisy},t}$, $w_{\text{clean},t}$, and $w_{\text{2D},t}$ be their corresponding downsampled frame weights.
Here $\phi_{\text{noisy}}$ and $\phi_{\text{clean}}$ are linear projections for 3D latent tokens, and $\phi_{\text{2D}}$ is a lightweight MLP for 2D tokens.
A modality is disabled by setting its weight to zero; the same mechanism also supports condition dropout and CFG \cite{ho2022classifier}. We fuse them by a weighted sum:

\vspace{-3mm}
\begin{equation}
\begin{aligned}
\mathbf{c}_t
&=
w_{\text{noisy},t}\,\phi_{\text{noisy}}(\mathbf{y}^{\text{noisy}}_t)
+
w_{\text{clean},t}\,\phi_{\text{clean}}(\mathbf{y}^{\text{clean}}_t) +
w_{\text{2D},t}\,\phi_{\text{2D}}(\mathbf{y}^{\text{2D}}_t),
\qquad t\in[1{:}T'].
\end{aligned}
\label{eq:weighted_fusion}
\end{equation}

\vspace{-3mm}
\paragraph{Condition injection}
We inject time aligned conditions through AdaLN modulation and residual gates, using the token $\mathbf{c}_t$ as the conditioning signal for each block.
Text is injected via cross attention: we extract token level features from a frozen CLIP text encoder, project them with MLP, and gate the cross-attention contribution to remain compatible with CFG \cite{ho2022classifier}.


\vspace{-1mm}
\subsection{Latent Motion Quality Discriminator (LMQD)}
\vspace{-1mm}

Regression losses in flow matching do not fully capture perceptual motion quality.
We draw inspiration from PHC~\cite{luo2023perpetual} in using adversarial supervision to improve motion realism, but instead apply the discriminator in the autoencoder latent space rather than on handcrafted kinematic features (joint positions, accelerations, power).
Operating in latent space offers two advantages: (i) gradients flow directly to the generator without passing through the decoder, and (ii) the learned latent captures more intrinsic motion structure than kinematic descriptors.
An ablation over five discriminator feature representations (Table~\ref{tab:ablation_lmqd_features}) confirms that latent-space discrimination achieves the best overall quality.
LMQD enforces a sequence-level distribution constraint and reduces skating, jitter, and penetration.

\vspace{-3mm}
\paragraph{Real and fake latents.}
The discriminator $D$ operates on whole latent sequences.
Real samples are normalized GT latents $\mathbf{z}_0^{\text{norm}}$.
Fake samples are built from the flow matching path.
We sample $(t, x_0, x_1)$ from the transport, obtain $x_t$, predict velocity $v_\theta(x_t,t,\mathcal{C})$, and flatten the sequence before feeding it to $D$..
We form a denoised latent prediction as
\vspace{-1mm}
\begin{equation}
\hat{x}_0 = x_0 + v_\theta(x_t,t,\mathcal{C}).
\end{equation}

\vspace{-4mm}
\paragraph{Joint training with alternating updates.}
We train CleanMDM and LMQD jointly with alternating optimization, following the standard practice used in CycleGAN~\cite{CycleGAN2017,isola2017image} training codebases.
At each iteration, we first freeze $D$ and update the generator with a flow-matching objective plus an adversarial term, where $\mathcal{L}_{\text{adv}}$ encourages $D(\hat{x}_0)$ to be classified as real; we then freeze the generator and update $D$ to separate $\hat{x}_0$ from $\mathbf{z}_0^{\text{norm}}$.
\vspace{-1mm}
\begin{equation}
\mathcal{L}_G = \mathcal{L}_{\text{flow}} + \lambda_{\text{adv}}\,\mathcal{L}_{\text{adv}},
\end{equation}

\vspace{-4mm}
\subsection{Auto Cleanup and Inbetweening in One Model}
\vspace{-1mm}
CleanMDM unifies mask-guided inbetweening and blind auto cleanup because animation repair often alternates between the two: an animator may correct corrupted poses, lock keyframes, and then inbetween the damaged span.
Using separate cleanup and inbetweening models would form a cascade in which the first stage can pass drift or over-smoothed poses to the second stage, while reliable regions may be rewritten twice.
We instead train one model with a shared condition interface.
For half of each training batch, we remove explicit interval supervision by marking all frames as clean in the corruption indicator, forcing the model to perform blind auto cleanup from the noisy draft.
For the other half, we keep the interval annotation and train mask-guided cleanup.
At inference, users may provide no interval annotation for automatic cleanup, provide corrupted intervals for inbetween-style repair, or additionally supply sparse 2D/3D keyframes and text prompts.
All cases use the same condition dictionary, and each modality is enabled or disabled by its confidence weight.




\vspace{-1mm}
\subsection{Mesh-Aware Contact Projection}

\vspace{-1mm}
Motion sequences produced by animation pipelines or generative models often contain artifacts such as foot sliding, floating, and ground penetration.
Manual cleanup is typically performed frame by frame.
Classic CCD-based IK operates frame-locally without temporal context, while physics-driven correction (e.g., PHC~\cite{luo2023perpetual}) may introduce motion distortion when strictly tracking input keypoints.
We therefore propose Mesh-Aware Contact Projection, a test-time optimization that refines the entire sequence under explicit ground contact constraints.
Let the environment be represented by a height field $g_{\psi}(x,z)$ that returns the ground height at horizontal location $(x,z)$.
This makes the formulation terrain-agnostic: a horizontal plane is the special case $g_{\psi}(x,z)=h_g$, where the scalar height $h_g$ can be optimized jointly with the SMPL parameters.

Since standard SMPL foot keypoints may lie inside the mesh, we sample 54 vertices on each foot sole to better approximate the contact surface.
Contact states for heel/toe are provided by a pre-trained estimator~\cite{mourot2022underpressure} or upstream predictions.
We solve a sequence-level optimization with a composite objective, the full definitions of which are provided in the Appendix~\ref{app:ik_details}, ~\ref{app:ik_table}.
\vspace{-1mm}
\begin{equation}
    \mathcal{L} = \lambda_{\text{recon}}\mathcal{L}_{\text{recon}} + \lambda_{\text{contact}}\mathcal{L}_{\text{contact}} + \lambda_{\text{reg}}\mathcal{L}_{\text{reg}}
\end{equation}

\vspace{-3mm}
\section{Experiment}
\label{sec:experiment}

\begin{table*}[t]
\centering
\caption{
\textbf{Comparison to prior work under three evaluation settings.}
We report results for auto cleanup (auto cleanup from a noisy draft, no interval annotation),
inbetween (mask guided interval completion),
and multimodal cleanup (interval completion with additional 2D/3D keyframes and text).
MPJPE(mm) reports global error, with local MPJPE in parentheses (root effects removed).
Skate denotes foot skating ratio; Acc and Jitter measure second- and third-order temporal errors; Pen.F(\%) and Pen.D(mm) denote ground penetration frequency and distance.
}
\footnotesize
\setlength{\tabcolsep}{2.4pt}
\renewcommand{\arraystretch}{1.05}
\resizebox{\textwidth}{!}{%

\begin{tabular}{@{}>{\raggedright\arraybackslash}m{1.4cm} l c c c c c c c c c c c@{}}
\toprule
\shortstack[l]{Setting} & \shortstack[l]{Method} &
MPJPE(mm)$\downarrow$ & Skate$\downarrow$ & Acc Err$\downarrow$ &
Pen.F(\%)$\downarrow$ & Pen.D(mm)$\downarrow$ & Jitter$\downarrow$ &
KF Err(mm)$\downarrow$ & FID$\downarrow$ & Div$\rightarrow$ & M2M Score$\uparrow$ & M2M R@3$\uparrow$ \\
\midrule \midrule

\multirow{3}{*}{Auto cleanup}
& GT & 0.00 & 0.13 & 0.00 & 0.00 & 0.00 & 3.73 & 0.00 & 0.00 & 9.40 & 1.00 & 100\% \\
\cline{2-13}
& StableMotion & 59.36 & 0.20 & 2.24 & 0.00 & 1.25 & 7.90 & \ding{55} & 0.12 & 7.09 & 0.95 & 90.72\% \\
& \cellcolor{ourblue}CleanMDM
  & \cellcolor{ourblue}\textbf{18.25(8.84)}
  & \cellcolor{ourblue}0.18
  & \cellcolor{ourblue}2.04
  & \cellcolor{ourblue}0.03
  & \cellcolor{ourblue}0.29
  & \cellcolor{ourblue}6.28
  & \cellcolor{ourblue}\ding{55}
  & \cellcolor{ourblue}\textbf{0.12}
  & \cellcolor{ourblue}\textbf{9.17}
  & \cellcolor{ourblue}\textbf{0.99}
  & \cellcolor{ourblue}\textbf{99.92\%} \\
& \cellcolor{ourblue}CleanMDM + MACP
  & \cellcolor{ourblue}45.15(19.91)
  & \cellcolor{ourblue}\textbf{0.14}
  & \cellcolor{ourblue}\textbf{1.88}
  & \cellcolor{ourblue}\textbf{0.00}
  & \cellcolor{ourblue}\textbf{0.00}
  & \cellcolor{ourblue}\textbf{3.74}
  & \cellcolor{ourblue}\ding{55}
  & \cellcolor{ourblue}0.24
  & \cellcolor{ourblue}9.16
  & \cellcolor{ourblue}0.96
  & \cellcolor{ourblue}98.73\% \\
\midrule

\multirow{5}{*}{Inbetween}
& GT & 0.00 & 0.12 & 0.00 & 0.00 & 0.00 & 3.15 & 0.00 & 0.00 & 9.44 & 1.00 & 100\% \\
\cline{2-13}
& ACMDM & 54.4 & 0.21 & 4.13 & 0.00 & 0.54 & 10.71 & 21.9 & 0.39 & 9.65 & 0.95 & 95.11\% \\
& CondMDI & 53.5 & 0.42 & 7.59 & 0.00 & 0.01 & 33.77 & 31.8 & 0.14 & 10.27 & 0.98 & 91.53\% \\
& \cellcolor{ourblue}\textbf{CleanMDM}
  & \cellcolor{ourblue}\textbf{42.84(29.28)}
  & \cellcolor{ourblue}0.18
  & \cellcolor{ourblue}\textbf{2.21}
  & \cellcolor{ourblue}0.15
  & \cellcolor{ourblue}1.04
  & \cellcolor{ourblue}6.23
  & \cellcolor{ourblue}\textbf{14.8}
  & \cellcolor{ourblue}\textbf{0.11}
  & \cellcolor{ourblue}\textbf{9.04}
  & \cellcolor{ourblue}\textbf{0.97}
  & \cellcolor{ourblue}\textbf{96.47\%} \\
& \cellcolor{ourblue}\textbf{CleanMDM + MACP}
  & \cellcolor{ourblue}71.86(37.14)
  & \cellcolor{ourblue}\textbf{0.12}
  & \cellcolor{ourblue}2.27
  & \cellcolor{ourblue}\textbf{0.00}
  & \cellcolor{ourblue}\textbf{0.00}
  & \cellcolor{ourblue}\textbf{4.24}
  & \cellcolor{ourblue}51.44
  & \cellcolor{ourblue}0.19
  & \cellcolor{ourblue}9.02
  & \cellcolor{ourblue}0.93
  & \cellcolor{ourblue}92.30\% \\
\midrule

\multirow{3}{*}{\shortstack[l]{Multimodal\\cleanup}}
& GT & 0.00 & 0.13 & 0.00 & 0.00 & 0.00 & 3.73 & 0.00 & 0.00 & 9.40 & 1.00 & 100\% \\
\cline{2-13}
& GenMo & 646.4 & 0.02 & 2.05 & 7.71 & 2.85 & \textbf{2.15} & 633.2 & 11.94 & 7.50 & 0.64 & 70\% \\
& \cellcolor{ourblue}\textbf{CleanMDM}
  & \cellcolor{ourblue}\textbf{21.95(15.73)}
  & \cellcolor{ourblue}0.18
  & \cellcolor{ourblue}\textbf{1.98}
  & \cellcolor{ourblue}0.06
  & \cellcolor{ourblue}0.76
  & \cellcolor{ourblue}5.54
  & \cellcolor{ourblue}\textbf{11.4}
  & \cellcolor{ourblue}\textbf{0.11}
  & \cellcolor{ourblue}9.23
  & \cellcolor{ourblue}\textbf{0.99}
  & \cellcolor{ourblue}\textbf{99.82\%} \\
& \cellcolor{ourblue}\textbf{CleanMDM + MACP}
  & \cellcolor{ourblue}50.91(24.77)
  & \cellcolor{ourblue}\textbf{0.14}
  & \cellcolor{ourblue}2.00
  & \cellcolor{ourblue}\textbf{0.00}
  & \cellcolor{ourblue}\textbf{0.00}
  & \cellcolor{ourblue}3.75
  & \cellcolor{ourblue}45.46
  & \cellcolor{ourblue}0.26
  & \cellcolor{ourblue}\textbf{9.24}
  & \cellcolor{ourblue}0.96
  & \cellcolor{ourblue}98.62\% \\
\bottomrule
\end{tabular}
}
\label{tab:baseline}
\vspace{-5mm}
\end{table*}

\vspace{-2mm}
\subsection{Experimental Setup}
\vspace{-2mm}
We build training clips from AMASS~\cite{AMASS:ICCV:2019} and MotionLLaMA~\cite{ling2024versatilemotion}, converting each sequence to a unified SMPL~\cite{loper2023smpl} 22-joint skeleton and absolute joint-position stream.
Each clip is normalized by centering the first-frame root on the horizontal plane and standardizing joint coordinates.
To emulate video-mocap degradations, we inject localized artifacts into random temporal intervals of clean clips, following the protocol in StableMotion~\cite{mu2025stablemotion} for part of the noise design.
The artifact families include joint jittering, foot sliding, horizontal drifting, and vertical floating.
We record a per-frame defect label $\boldsymbol{\ell}\in\mathbb{N}^T$, where $\boldsymbol{\ell}[t]=0$ denotes a clean frame and non-zero values indicate corrupted frames.
Full data filtering, corruption construction, and localization details are provided in Appendix~\ref{sec:corruption_appendix}.

We use flow matching~\cite{lipman2022flow} with linear transport paths and velocity prediction.
Sampling uses an Euler ODE solver with $50$ steps.
The generator learning rate is $4\times 10^{-4}$ and the discriminator learning rate is $5\times 10^{-5}$, both with AdamW~\cite{loshchilov2017fixing}.
We set $\lambda_{\text{adv}}=0.01$, condition dropout probability $0.1$, and EMA decay $0.9999$.
The generator and discriminator are trained with alternating updates.
Our model is trained on $2$ H20 GPUs with a batch size of $128$ for $600$ epochs (${\sim}16$ hours) and tested on 1 NVIDIA V100 GPU. Metric definitions are provided in Appendix~\ref{sec:metrics}.


\vspace{-2mm}
\subsection{Ablation Settings}
\vspace{-2mm}
We denote CleanMDM variants by C0--C6 for compact reporting.
All variants share the same latent flow matching backbone and the same input noisy draft; they differ only in enabled conditions and post-processing.
Table~\ref{tab:config} summarizes the configuration of each variant.

\vspace{-2mm}
\subsection{Comparison}
\label{sec:sota}
\vspace{-2mm}
\paragraph{Evaluation settings.}
Table~\ref{tab:baseline} compares CleanMDM with representative restoration and conditional generation baselines under three settings.
\textbf{(i) Auto cleanup} evaluates auto cleanup from a single noisy 3D draft without any interval annotation.
To keep the comparison fair, multimodal baselines are restricted to the same noisy input, and we provide a default, non-informative prompt (``a person is doing something'').
\textbf{(ii) Inbetween} evaluates mask guided cleanup where the corrupted interval is removed and the method must inbetween the missing segment.
We follow the standard split used by CondMDI, corrupt motions in its test set, and apply the same interval masking for all methods.
Text-based methods receive the caption associated with each sequence.
\textbf{(iii) Multimodal cleanup} evaluates the full production oriented setup, where the corrupted interval is masked and additional user cues are available (sparse 3D keyframes, sparse 2D keyframes, and text).
Since no existing cleanup method supports comparable multimodal inputs, we compare against GenMo~\cite{li2025genmo} --- a multimodal generation model that accepts 2D/3D keyframes and text --- under the same masked-interval completion setting.
We run GenMo with the same inputs; when required, we also provide its camera-related inputs following its inference protocol.
Note that GenMo was designed for motion generation rather than cleanup; this comparison therefore highlights the gap between generation-oriented and cleanup-oriented approaches when applied to the same repair task.

\begin{table*}[t]
\centering
\caption{\textbf{Main ablation on multimodal conditions and regularizers.}
C0--C6 differ only in the enabled conditions and post-processing (see Table~\ref{tab:config}).}
\label{tab:ablation_main}
\footnotesize
\setlength{\tabcolsep}{3.0pt}
\renewcommand{\arraystretch}{1.08}
\begin{tabular}{@{}lccccccccc@{}}
\toprule
Method & MPJPE$\downarrow$ & Skate$\downarrow$ & Acc$\downarrow$ &
Pen.F$\downarrow$ & Pen.D$\downarrow$ & Jitter$\downarrow$ &
KF Err$\downarrow$ & FID$\downarrow$ & Div$\rightarrow$ \\
\midrule \midrule
GT & 0.00 & 0.13 & 0.00 & 0.00 & 0.00 & 3.73 & 0.00 & 0.00 & 9.40 \\
\midrule
C0 & 35.45 & 0.19 & 2.12 & 0.07 & 0.95 & 5.73 & 12.7 & 0.09 & 8.04 \\
C1 & 34.33 & 0.21 & 2.27 & 0.08 & 0.84 & 6.26 & 13.2 & \textbf{0.07} & 7.95\\
C2 & 27.45 & 0.20 & 2.17 & 0.07 & 0.89 & 6.12 & 13.3 & \textbf{0.07} & 8.05 \\
C3 & 27.32 & 0.20 & 2.19 & 0.06 & 0.78 & 6.17 & 13.0 & 0.08 & 8.06 \\
C4 & 28.87 & 0.19 & 2.07 & 0.05 & 0.72 & 5.74 & 12.2 & 0.13 & 9.17 \\
\rowcolor{ourblue}
\textbf{C5 (CleanMDM)} & \textbf{21.95} & 0.18 & \textbf{1.98} & 0.06 & 0.76 & 5.54 & \textbf{11.4} & 0.11 & 9.23 \\
\rowcolor{ourblue}
\textbf{C6 (CleanMDM + MACP)} & 50.91 & \textbf{0.14} & 2.00 & \textbf{0.00} & \textbf{0.00} & \textbf{3.75} & 45.46 & 0.26 & \textbf{9.24} \\
\bottomrule
\end{tabular}
\vspace{-5mm}
\end{table*}

\vspace{-3mm}
\paragraph{Main results.}
Across all three settings, CleanMDM achieves the strongest overall motion quality.
In \textbf{auto cleanup}, CleanMDM substantially improves reconstruction fidelity and distributional realism over StableMotion, while simultaneously reducing contact-related artifacts (skating and penetration) and temporal noise (acceleration error and jitter).
In \textbf{inbetween}, CleanMDM outperforms ACMDM and CondMDI on both geometric accuracy (MPJPE and keyframe error) and kinematic quality, indicating that the same backbone can handle interval completion and denoising within a unified formulation.
In the most challenging \textbf{multimodal cleanup} setting, CleanMDM substantially outperforms GenMo (21.95 vs.\ 646.4 MPJPE), consistent with our expectation that a model designed for generation rather than cleanup does not explicitly preserve the input draft, resulting in lower reconstruction fidelity.
Fig.~\ref{fig:qual_baseline} (~\ref{sec:qualitative}) visualizes representative examples under the same three settings, highlighting improved contact quality (e.g., reduced foot sliding) and stronger keyframe adherence compared with task matched baselines.


\vspace{-3mm}
\paragraph{Real-world validation.}
Beyond synthetic corruption, we evaluate CleanMDM on real video-mocap outputs from animation, monocular video (HAA500, IDEA400, Kungfu) and multi-view video (HuMMan) sources, which contain artifacts beyond our training corruption types.
Table~\ref{tab:public_datasets_main} moves two representative public-dataset results (IDEA400 and HuMMan) into the main paper.
On KungFu, CleanMDM cuts MPJPE roughly in half compared with StableMotion (69.4 vs.\ 127.8), while StableMotion attains a marginally lower jitter (9.2 vs.\ 12.0) by over-smoothing the motion in a way that alters its action identity; Appendix Fig.~\ref{fig:kungfu_unfolded} visualizes this trade-off on a \emph{Tornado Kick} clip, where StableMotion exhibits visible foot floating and a weakened rotational signature relative to the real-world input.
Full public-dataset evaluations and real-world visualizations are provided in Appendix~\ref{sec:public} (Table~\ref{tab:public_datasets}, Figs.~\ref{fig:kungfu_unfolded}--\ref{fig:kungfu_basketball}).
These results confirm that CleanMDM's motion prior generalizes to real-world degradations, not just the synthetic artifacts used during training.


\vspace{-3mm}
\paragraph{Effect of Mesh-Aware Contact Projection.}
C6 applies our Mesh-Aware Contact Projection as a post-process.
While global MPJPE can increase (due to sequence-level adjustments of root/ground alignment), C6 consistently reduces artifacts, including skating, penetration frequency/distance, and jitter, approaching the statistics of the ground truth.
This behavior matches the role of Mesh-Aware Contact Projection as a projection step: it enforces hard contact consistency and temporal coherence, which are only weakly captured by regression-style training objectives.
Therefore, CleanMDM provides a strong generative cleanup prior, while C6 further improves physical plausibility when strict contact quality is required. Fig.~\ref{fig:ik_qual} shows a representative example on GVHMR outputs, where Mesh-Aware Contact Projection removes foot sliding and corrects floating artifacts.

\begin{table*}[t]
\centering
\begin{minipage}[t]{0.47\textwidth}
\centering
\caption{\textbf{C0--C6 configurations.} All variants share the same backbone.}
\label{tab:config}
\footnotesize
\setlength{\tabcolsep}{2.4pt}
\renewcommand{\arraystretch}{1.06}
\resizebox{\linewidth}{!}{%
\begin{tabular}{@{}lccccc@{}}
\toprule
Config & Text & 2DKF & LMQD & 3DKF & MACP \\
\midrule
C0 & & & & & \\
C1 & \checkmark & & & & \\
C2 & & \checkmark & & & \\
C3 & \checkmark & \checkmark & & & \\
C4 & \checkmark & \checkmark & \checkmark & & \\
\rowcolor{ourblue} \textbf{C5} & \checkmark & \checkmark & \checkmark & \checkmark & \\
\rowcolor{ourblue} \textbf{C6} & \checkmark & \checkmark & \checkmark & \checkmark & \checkmark \\
\bottomrule
\end{tabular}
}
\end{minipage}
\hfill
\begin{minipage}[t]{0.49\textwidth}
\centering
\caption{\textbf{Selected real-world public-dataset results.}
Full results are in Appendix~\ref{sec:public}, Table~\ref{tab:public_datasets}; MPJPE measures drift to the real-world input.}
\label{tab:public_datasets_main}
\footnotesize
\setlength{\tabcolsep}{2.0pt}
\renewcommand{\arraystretch}{1.06}
\resizebox{\linewidth}{!}{%
\begin{tabular}{@{}l l c c c c@{}}
\toprule
Dataset & Method & \shortstack{MPJPE\\(mm)$\downarrow$} & Jitter$\downarrow$ & \shortstack{Slide\\R$\downarrow$} & \shortstack{Slide\\A$\downarrow$} \\
\midrule \midrule
\multirow{3}{*}{IDEA400}
& Input & 0.00 & 5.43 & 0.30 & 0.06 \\
& StableMotion & 40.10 & 5.10 & 0.45 & 0.14 \\
& \cellcolor{ourblue}\textbf{CleanMDM}
    & \cellcolor{ourblue}\textbf{15.61}
    & \cellcolor{ourblue}\textbf{4.94}
    & \cellcolor{ourblue}\textbf{0.25}
    & \cellcolor{ourblue}\textbf{0.05} \\
\midrule
\multirow{3}{*}{HuMMan}
& Input & 0.00 & 10.30 & 0.41 & 0.08 \\
& StableMotion & \textbf{84.02} & 10.17 & 0.36 & 0.07 \\
& \cellcolor{ourblue}\textbf{CleanMDM}
    & \cellcolor{ourblue}84.72
    & \cellcolor{ourblue}\textbf{7.80}
    & \cellcolor{ourblue}\textbf{0.32}
    & \cellcolor{ourblue}\textbf{0.06} \\
\bottomrule
\end{tabular}
}
\end{minipage}
\vspace{-5mm}
\end{table*}

\vspace{-2mm}
\subsection{Ablation Studies}
\label{sec:ablation}
\vspace{-2mm}
\paragraph{Adding modalities improves controllable cleanup.}
Table~\ref{tab:ablation_main} shows that motion cleanup benefits from progressively adding complementary modalities.
Starting from C0 (noisy 3D draft only), adding text (C1) provides coarse semantic and style guidance but yields limited improvements on geometric metrics.
Adding sparse 2D keyframes (C2) consistently improves accuracy and keyframe adherence, suggesting that 2D cues act as an effective geometric scaffold even without explicit 2D-to-3D lifting.
Combining 2D and text (C3) further improves the overall balance between fidelity and controllability.

Notably, the strongest gains appear when CleanMDM additionally leverages sparse 3D keyframes, which anchor the reconstruction inside corrupted intervals and reduce ambiguity.
This trend supports the central motivation of multimodal cleanup: different cues compensate for each other under partial corruption, and a single model can exploit whichever signals are available at test time.

\vspace{-3mm}
\paragraph{LMQD improves distribution-level motion quality.}
Comparing C3 and C4 isolates the effect of LMQD.
Although MPJPE may not monotonically decrease, LMQD improves contact and temporal quality metrics (e.g., reduced penetration distance and jitter) and increases diversity relative to ground truth motion, indicating a better match to the motion manifold beyond point-wise regression.
This supports our design choice of imposing a sequence-level adversarial regularizer in latent space: it is lightweight, compatible with flow matching, and directly targets artifacts that are perceptually salient yet poorly captured by deterministic reconstruction losses.

\vspace{-3mm}
\paragraph{Why latent fusion matters.}
Replacing autoencoder-based 3D keyframe encoding with direct MLP projection in joint space (C5 (MLP), Appendix~\ref{sec:ablation_latent_vs_mlp} Table~\ref{tab:latent_vs_mlp}) degrades temporal quality dramatically (Jitter: $71.02$ vs.\ $5.54$; FID: $2.99$ vs.\ $0.11$).
We attribute the gap to the autoencoder's temporal downsampling bottleneck, which provides a normalized coordinate system and temporal-coherence prior that makes weighted fusion well-posed across heterogeneous signals; raw-space MLP encoding lacks this structure and creates a modality gap with the AE-encoded noisy draft.
Moreover, multimodal training itself improves the motion prior: C5 outperforms C0 even when no extra conditions are given at inference (Appendix~\ref{sec:ablation_multimodal_denoise}, Table~\ref{tab:multimodal_denoise}).

\vspace{-1mm}
\section{Conclusions, Limitations and Future Work}
\label{sec:conclusion}
\vspace{-2mm}
We presented CleanMDM, a unified framework for motion cleanup that formulates repair and inbetweening as masked conditional generation with plug-and-play multimodal conditions, including noisy 3D drafts, sparse 2D/3D keyframes, and text prompts. CleanMDM fuses heterogeneous geometric cues in a shared latent representation with frame-wise weighting, enabling the model to preserve reliable observations while regenerating corrupted regions under incomplete or mixed constraints. To better address perceptual artifacts that are weakly captured by regression-style objectives, we introduced LMQD, and further applied Mesh-Aware Contact Projection to enforce contact consistency and reduce skating and penetration. Experiments across auto repair, inbetweening, and multimodal cleanup demonstrate consistent improvements over prior baselines, with particularly strong gains on contact and temporal quality metrics.

A current limitation is that our evaluation relies on controlled corruption and a simplified ground model during Mesh-Aware Contact Projection. In future work, we plan to incorporate more realistic video mocap degradations, stronger learned contact priors, and additional control modalities and interactive authoring workflows for animation pipelines.

\bibliographystyle{abbrvnat}
\bibliography{sample-base}

\appendix

\newpage
\section{Supplementary Material}

\subsection{Qualitative Comparisons}
\label{sec:qualitative}

\begin{figure*}[t]
    \centering
    \includegraphics[width=0.95\linewidth]{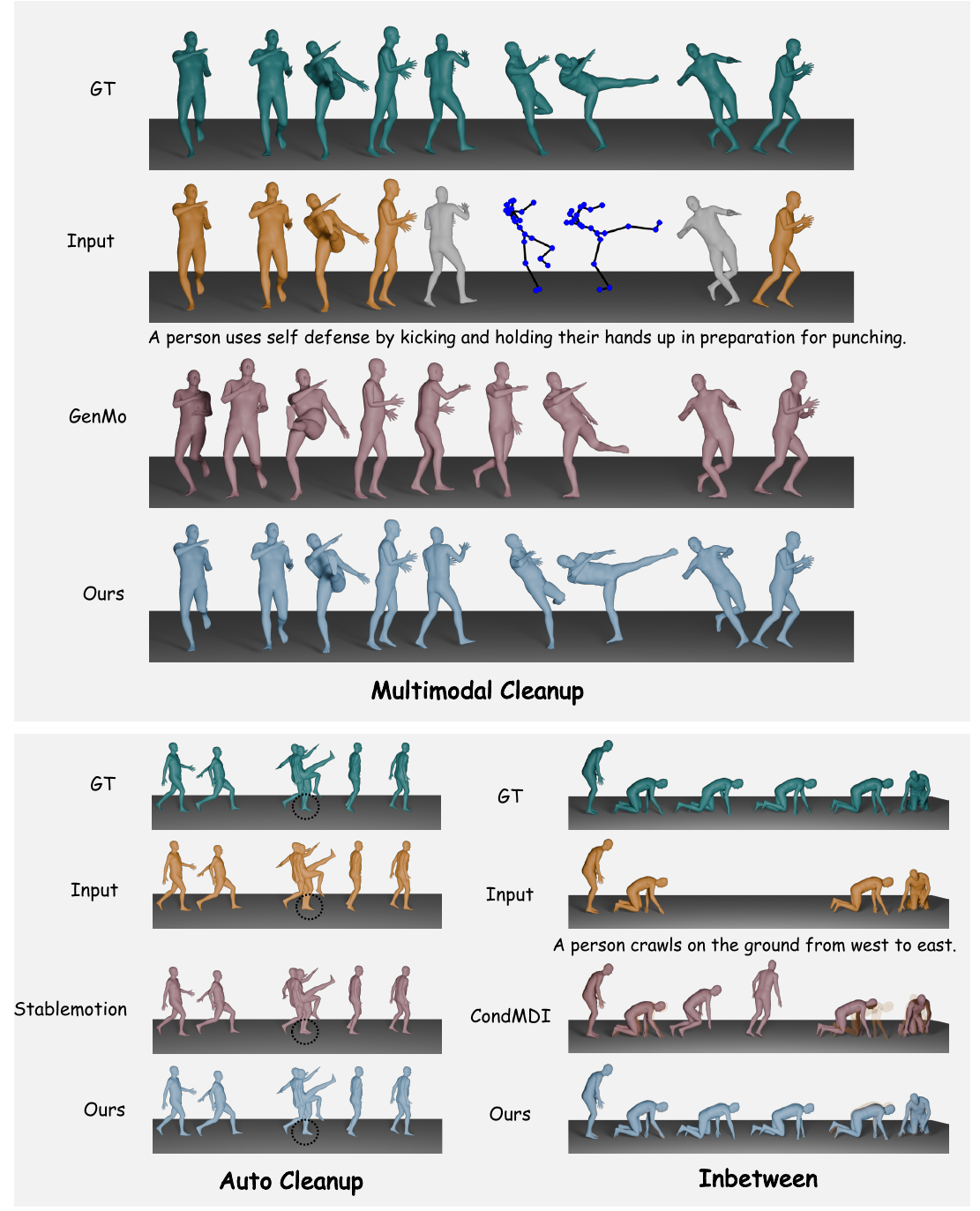}
    \caption{\textbf{Qualitative comparisons with task-matched baselines.}
Top: \textbf{multimodal cleanup} compared with GenMo.
The input consists of an unmasked noisy draft (yellow), sparse 3D keyframes (gray bodies), sparse 2D keyframes (blue skeleton), and a text prompt.
Bottom-left: \textbf{auto cleanup} compared with StableMotion; CleanMDM corrects foot sliding (circled) while preserving the remaining motion.
Bottom-right: \textbf{inbetween} compared with CondMDI; CleanMDM better satisfies the 3D keyframe constraints (yellow translucent overlays) and yields smoother transitions.
Rows show \textit{GT}, \textit{Input}, \textit{Baseline}, and \textit{Ours}.}
    \label{fig:qual_baseline}
\end{figure*}

Figure~\ref{fig:qual_baseline} provides side-by-side qualitative comparisons 
under all three evaluation settings. In multimodal cleanup (top), CleanMDM 
produces motion that closely follows the provided keyframes and text while 
maintaining natural contact, whereas GenMo drifts from the input draft. 
In auto cleanup (bottom-left), CleanMDM corrects localized foot sliding 
while preserving the reliable upper-body motion. In inbetweening (bottom-right), 
CleanMDM achieves smoother transitions and better keyframe adherence than CondMDI.

\subsection{Data}
\label{sec:corruption_appendix}

To emulate video mocap degradations, we synthesize corrupted motions by perturbing SMPL~\cite{loper2023smpl} joint rotations and root translations, following the artifact injection protocol in StableMotion~\cite{mu2025stablemotion} for part of the noise design.
We apply a random subset of artifact types to a random temporal interval $[s,e)$, including joint jittering, foot sliding, horizontal drifting, and vertical floating.
To ensure that corruptions remain localized and do not alter the clean segment outside the interval, we inject translation based artifacts in the velocity domain and integrate them only within $[s,e)$, while enforcing zero net displacement over the interval (i.e., the integrated root translation offset satisfies $\Delta\mathbf{t}_{e}-\Delta\mathbf{t}_{s}=\mathbf{0}$), so that frames after $e$ keep the original root trajectory unchanged.
We record a per-frame defect label $\mathrm{det\_mask}\in\mathbb{N}^T$, where $\mathrm{det\_mask}[t]=0$ denotes a clean frame and non-zero values indicate corrupted frames.

\paragraph{Data Sources}
We build our training clips from AMASS~\cite{AMASS:ICCV:2019} and MotionLLaMA~\cite{ling2024versatilemotion}. MotionLLaMA originally contains approximately 48K motion-text pairs; however, part of this collection is noisy, lacks reliable text annotations, or comes from auxiliary sources that are less suitable for our setting. We therefore discard all MotionGV~\cite{fan2025go} pseudo-labeled data, samples without text annotations, and most low-quality subsets from Motion Union, and retain only the subsets marked in Table~\ref{tab:motionllama_subsets}. After filtering, about 28K MotionLLaMA clips remain for training. The final dataset is split into training, validation, and test sets with a ratio of 80:15:5.

\paragraph{Absolute Joint Positions}
We represent motion in absolute joint positions $\mathbf{X}_{1:T}\in\mathbb{R}^{T\times J\times 3}$, following ACMDM~\cite{meng2025absolute}.
This choice is motivated by the cleanup interface.
First, absolute positions respond directly to spatial control signals: sparse 3D keyframes are already in the same coordinate system, and 2D keypoints are obtained through camera projection.
Second, absolute coordinates avoid reconstructing global motion by accumulating local offsets, which can propagate small per-frame errors into long-horizon drift.
Third, recent transformer and diffusion backbones~\cite{peebles2023scalable} can internalize many kinematic regularities from data, so a concise representation reduces modeling complexity without sacrificing fidelity.
Finally, contact detection and inference-time IK operate naturally in the absolute domain, which facilitates stable enforcement of foot contact and ground constraints.

\paragraph{Artifact types.}
We simulate four common families: (i) \textbf{jittering} (high-frequency rotational noise on a subset of joints),
(ii) \textbf{floating} (vertical root drift),
(iii) \textbf{foot sliding} (spurious horizontal root motion causing skating),
and (iv) \textbf{drifting} (slow horizontal root drift).
The corrupted interval length is sampled to be localized (longer during training, shorter at test time).

\paragraph{Localized corruption via integrated, zero-displacement perturbations.}
For translation-based artifacts (foot sliding, drifting, floating), we inject noise in the \emph{velocity} domain and integrate it \emph{only} inside $[s,e)$, while enforcing approximately zero net displacement so that frames after $e$ remain unchanged.
Concretely, we generate a velocity perturbation $\Delta\dot{\mathbf{t}}_{s:e}$, taper it with a smooth window whose endpoints are zero, and remove its (windowed) mean so that the integrated offset cancels out:
\begin{equation}
\Delta\dot{\mathbf{t}}'_\ell
=
\Delta\dot{\mathbf{t}}_\ell
-
w_\ell \cdot
\frac{\sum_{k=0}^{L-1}\Delta\dot{\mathbf{t}}_k}{\sum_{k=0}^{L-1} w_k+\epsilon},
\qquad
L=e-s,
\end{equation}
followed by integration to obtain $\Delta\mathbf{t}$ and applying $\mathbf{t}'_{s:e}=\mathbf{t}_{s:e}+\Delta\mathbf{t}$ while keeping $\mathbf{t}'_{t}=\mathbf{t}_{t}$ for $t\notin[s,e)$.
In practice we additionally enforce an exact endpoint correction to eliminate numerical drift.
This construction ensures the corruption is visually salient within the interval but does not affect the clean segment after it, enabling fair side-by-side evaluation.

\paragraph{Relation to StableMotion.}
Our artifact families and part of the injection protocol are adapted from StableMotion.
A key difference is how we apply translation-based corruptions.
Instead of directly perturbing root translations in position space (which can accumulate drift and contaminate frames outside the intended interval),
we inject translation noise in the \emph{velocity} domain and integrate it \emph{only} within $[s,e)$, enforcing (approximately) zero net displacement over the interval.
This construction ensures that the corrupted segment is visually salient while the clean motion before $s$ and after $e$ remains unchanged, enabling non-destructive cleanup evaluation with strictly preserved clean regions.

\begin{table}[t]
\centering
\small
\caption{MotionLLaMA-related subsets shown in the data manager.}
\label{tab:motionllama_subsets}
\setlength{\tabcolsep}{8pt}
\begin{tabular}{lll}
\toprule
Source Group & Dataset & Used \\
\midrule
Motion LLaMA & HumanSC3D  & \checkmark \\
Motion LLaMA & FineDance  & \checkmark \\
Motion LLaMA & Fit3D      & \checkmark \\
Motion LLaMA & Hi4D       & \checkmark \\
Motion LLaMA & InterHuman & \ding{55} \\
Motion LLaMA & InterX     & \checkmark \\
Motion LLaMA & TruMans    & \checkmark \\
\midrule
Motion Union & 100STYLE      & \checkmark \\
Motion Union & EgoBody       & \ding{55}  \\
Motion Union & Game Motion   & \ding{55}  \\
Motion Union & Platform      & \ding{55}  \\
Motion Union & Music         & \ding{55}  \\
Motion Union & KungFu        & \ding{55}  \\
Motion Union & IDEA400       & \ding{55}  \\
Motion Union & HuMMan        & \ding{55}  \\
Motion Union & HAA500        & \ding{55}  \\
Motion Union & Fitness       & \ding{55}  \\
Motion Union & Combat Motion & \ding{55}  \\
Motion Union & Animation     & \ding{55}  \\
\midrule
GVHMR Pseudo-label Data & MotionGV & \ding{55} \\
\bottomrule
\end{tabular}
\end{table}

\begin{table*}[t]
\centering
\caption{\textbf{Effect of Mesh-Aware Contact Projection on video-mocap outputs.}
We compare different post-processing strategies applied to GVHMR results on AIST++ and animation\_ref.
Our Mesh-Aware Contact Projection improves contact quality (lower skating and penetration) and maintains a high success ratio, indicating stable optimization behavior across datasets.}
\footnotesize
\setlength{\tabcolsep}{4.0pt}
\renewcommand{\arraystretch}{1.10}
\resizebox{\textwidth}{!}{%
\begin{tabular}{lccccc|ccccc}
\toprule
& \multicolumn{5}{c|}{AIST++ (128)} & \multicolumn{5}{c}{animation\_ref (20)} \\
\cmidrule(lr){2-6}\cmidrule(lr){7-11}
Method
& Skate$\downarrow$ & Jitter$\downarrow$ & Pen.F$\downarrow$ & Pen.D$\downarrow$ & Succ.$\uparrow$
& Skate$\downarrow$ & Jitter$\downarrow$ & Pen.F$\downarrow$ & Pen.D$\downarrow$ & Succ.$\uparrow$ \\
\midrule
GVHMR (CCD-IK) & 0.13 & \textbf{23.1513} & 14.20 & 48.46 & \textbf{1.00} & 0.2594 & 48.0547 & 17.53 & 40.41 & \textbf{1.00} \\
GVHMR (CCD-IK) + PHC & 0.0913 & 30.7293 & \textbf{0.00} & \textbf{0.95} & 0.867 & 0.0725 & \textbf{34.2244} & \textbf{0.00} & 1.31 & 0.35 \\
\rowcolor{ourblue} GVHMR + MACP (ours) & \textbf{0.056} & 27.3287 & 0.49 & 1.63 & \textbf{1.00} & \textbf{0.0583} & 63.5704 & \textbf{0.00} & \textbf{0.78} & \textbf{1.00} \\
\bottomrule
\end{tabular}
}
\label{tab:ik_fitting_eval}
\end{table*}

\subsection{Mesh-Aware Contact Projection: Additional Details}
\label{app:ik_details}

\noindent\textbf{General terrain formulation.}
Given an input SMPL sequence with body pose $\boldsymbol{\theta}^{\mathrm{in}}_{1:T}$ and root translation $\mathbf{r}^{\mathrm{in}}_{1:T}$, MACP optimizes the body pose $\boldsymbol{\theta}_{1:T}$, root translation $\mathbf{r}_{1:T}$, and, when needed, environment parameters $\psi$.
Let $\mathbf{v}^{s}_{t,i}=(x^{s}_{t,i},y^{s}_{t,i},z^{s}_{t,i})$ denote the $i$-th sampled sole vertex of foot $s\in\{L,R\}$ at frame $t$; we use 54 vertices per foot.
The environment is represented as a height field $g_{\psi}(x,z)$.
For any sole vertex, define its signed height residual relative to the environment as
\begin{equation}
    d^{s}_{t,i} =
    y^{s}_{t,i} - g_{\psi}\!\left(x^{s}_{t,i}, z^{s}_{t,i}\right).
\end{equation}
The horizontal ground case used in our main experiments is the special case $g_{\psi}(x,z)=h_g$, where the scalar $h_g$ is optimized together with the motion.
If a scanned, reconstructed, or procedural terrain is available, the same losses apply by replacing the constant $h_g$ with the corresponding height field.

\noindent\textbf{Objective.}
Let $c_{t,s}\in\{0,1\}$ be the contact label for foot $s$ at frame $t$, obtained from a pre-trained contact estimator~\cite{mourot2022underpressure} or from upstream predictions.
The contact labels are thresholded at 0.5 in our implementation.
Let $\rho(\cdot)$ map axis-angle joint rotations to the 6D rotation representation, and let $\Delta$ denote the first-order temporal difference.
MACP minimizes
\begin{align}
\mathcal{L}_{\text{MACP}}
=&\;
1\cdot\mathcal{L}_{\text{rot}}
+10^{-4}\cdot\mathcal{L}^{xz}_{\text{vel}}
+10^{-3}\cdot\mathcal{L}^{y}_{\text{vel}}
\nonumber\\
&+10^{-1}\cdot\mathcal{L}_{\text{height}}
+10^{-1}\cdot\mathcal{L}_{\text{pen}}
+10^{-1}\cdot\mathcal{L}_{\text{lock}}
\nonumber\\
&+10^{-2}\cdot\mathcal{L}^{\theta}_{\text{smooth}}
+10^{-4}\cdot\mathcal{L}^{r}_{\text{smooth}}.
\label{eq:macp_full_objective}
\end{align}
The reconstruction terms preserve the input motion:
\begin{align}
\mathcal{L}_{\text{rot}} &=
\frac{1}{T}\sum_{t=1}^{T}
\left\|\rho(\boldsymbol{\theta}_{t})-\rho(\boldsymbol{\theta}^{\mathrm{in}}_{t})\right\|_2^2,\\
\mathcal{L}^{xz}_{\text{vel}} &=
\frac{1}{|\Omega_{\bar c}|}\sum_{(t,s)\in\Omega_{\bar c}}
\left\|\Delta\mathbf{p}^{s,xz}_{t}-\Delta\mathbf{p}^{s,xz,\mathrm{in}}_{t}\right\|_1,\\
\mathcal{L}^{y}_{\text{vel}} &=
\frac{1}{|\Omega_{\bar c}|}\sum_{(t,s)\in\Omega_{\bar c}}
\left|\Delta p^{s,y}_{t}-\Delta p^{s,y,\mathrm{in}}_{t}\right|.
\end{align}
Here $\mathbf{p}^{s}_{t}$ denotes the ankle/toe foot joints and $\Omega_{\bar c}$ indexes non-contact foot frames.
The contact terms use the mesh sole vertices:
\begin{align}
\mathcal{L}_{\text{height}} &=
\frac{1}{|\Omega_c|}\sum_{(t,s)\in\Omega_c}\frac{1}{|S_s|}
\sum_{i\in S_s}\left(d^{s}_{t,i}\right)^2,\\
\mathcal{L}_{\text{pen}} &=
\frac{1}{|\Omega_{\bar c}|}\sum_{(t,s)\in\Omega_{\bar c}}\frac{1}{|S_s|}
\sum_{i\in S_s}\left[\max\left(0,-d^{s}_{t,i}\right)\right]^2,\\
\mathcal{L}_{\text{lock}} &=
\frac{1}{|\Omega_{cc}|}\sum_{(t,s)\in\Omega_{cc}}\frac{1}{|S_s|}
\sum_{i\in S_s}
\left\|\mathbf{v}^{s}_{t+1,i}-\mathbf{v}^{s}_{t,i}\right\|_2^2,
\end{align}
where $S_s$ is the sampled sole-vertex set for foot $s$, $\Omega_c$ indexes contact frames, and $\Omega_{cc}$ indexes consecutive contact frames.
Finally, temporal regularization suppresses jitter:
\begin{align}
\mathcal{L}^{\theta}_{\text{smooth}} &=
\frac{1}{T-1}\sum_{t=1}^{T-1}
\left\|\rho(\boldsymbol{\theta}_{t+1})-\rho(\boldsymbol{\theta}_{t})\right\|_2^2,\\
\mathcal{L}^{r}_{\text{smooth}} &=
\frac{1}{T-1}\sum_{t=1}^{T-1}
\left\|\mathbf{r}_{t+1}-\mathbf{r}_{t}\right\|_2^2.
\end{align}

\begin{table}[t]
\centering
\caption{\textbf{Loss weights used in Mesh-Aware Contact Projection.}
The table expands the implementation weights for Eq.~\ref{eq:macp_full_objective}.}
\label{tab:macp_loss_weights}
\footnotesize
\setlength{\tabcolsep}{5.5pt}
\begin{tabular}{lll}
\toprule
Loss Category & Loss Term & Weight \\
\midrule
Reconstruction & Rotation preservation & $1$ \\
Reconstruction & Horizontal velocity preservation (non-contact) & $10^{-4}$ \\
Reconstruction & Vertical velocity preservation (non-contact) & $10^{-3}$ \\
\midrule
Foot contact & Sole mesh height consistency & $10^{-1}$ \\
Foot contact & Sole mesh non-penetration & $10^{-1}$ \\
Foot contact & Zero velocity for contacting sole vertices & $10^{-1}$ \\
\midrule
Smoothness & Rotational smoothness & $10^{-2}$ \\
Smoothness & Translational smoothness & $10^{-4}$ \\
\bottomrule
\end{tabular}
\end{table}

\noindent\textbf{Optimization.}
We optimize the full sequence with Adam for 600 iterations using an initial learning rate of $10^{-2}$ and decay it by $0.1$ every 250 iterations.
The mesh-based terms are evaluated on the sampled SMPL sole vertices, while the non-contact velocity preservation terms use ankle/toe joints to preserve the original swing-foot motion.

\subsection{Additional Evaluation of Mesh-Aware Contact Projection}
\label{app:ik_table}

To further validate the effectiveness of our Mesh-Aware Contact Projection, we perform an auxiliary study on motions produced by a representative video-mocap system (GVHMR) under different post-processing strategies.
We evaluate on two datasets (AIST++ and animation\_ref) and report contact- and temporal-quality metrics, including Foot Skating Ratio, Jitter Error, Ground Penetration Frequency/Distance, together with a Success Ratio indicating whether the optimization converges to a feasible solution under our constraints.

\begin{figure*}[t]
    \centering
    \includegraphics[width=1\linewidth]{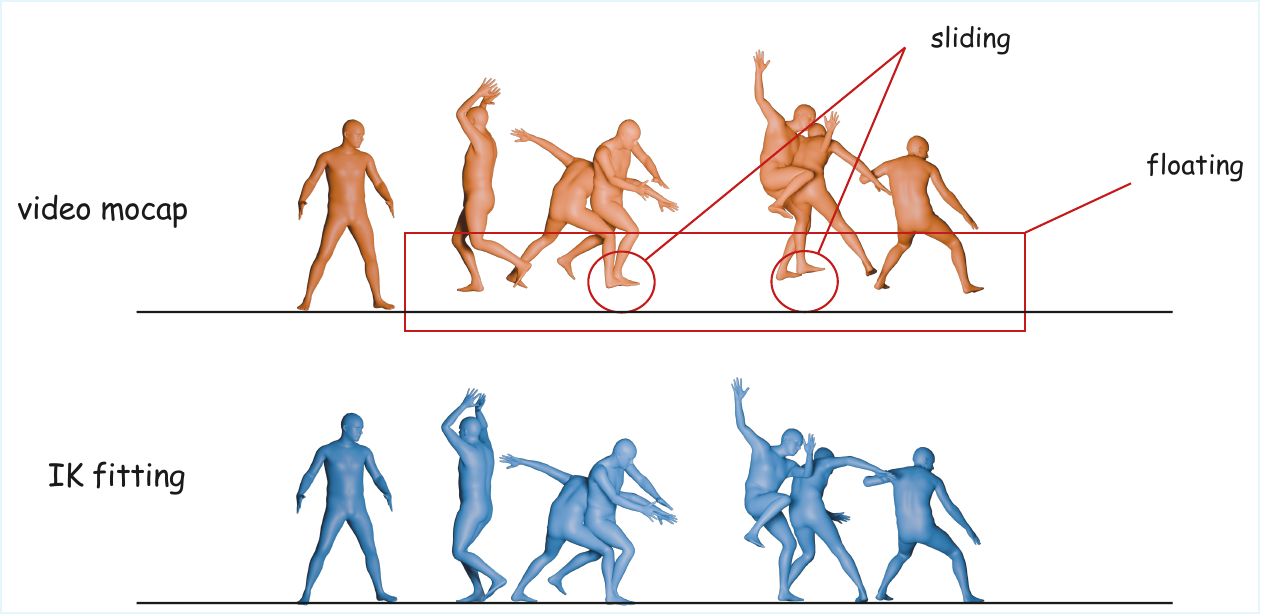}
    \caption{\textbf{Effect of test-time Mesh-Aware Contact Projection on video mocap outputs.}
Top: raw GVHMR \cite{shen2024world} results exhibit contact artifacts, including foot sliding and occasional floating (highlighted).
Bottom: applying our Mesh-Aware Contact Projection as a post-process enforces more consistent ground contact, reducing sliding and correcting floating while preserving the overall motion.}
    \label{fig:ik_qual}
\end{figure*}

Table~\ref{tab:ik_fitting_eval} shows that applying our Mesh-Aware Contact Projection consistently improves the kinematic plausibility of GVHMR outputs, substantially reducing foot skating and ground penetration compared to the standard CCD-IK post-processing used in GVHMR.
Figure~\ref{fig:ik_qual} visualizes the effect of Mesh-Aware Contact Projection 
on GVHMR outputs. The raw video-mocap results (top) exhibit visible foot sliding 
and occasional floating. After applying our contact projection (bottom), 
ground contact becomes consistent, sliding is eliminated, and floating is 
corrected, while the overall motion trajectory and style are preserved.
Compared to physics-based refinement (PHC) on top of CCD-IK, our method achieves a more reliable success ratio while still providing strong gains on penetration-related metrics.
Overall, this study supports the role of Mesh-Aware Contact Projection as an effective, lightweight projection step that enhances contact consistency for video-mocap motions, and explains why we adopt it as a plug-in post-process in our pipeline.

\subsection{Evaluation Metrics}
\label{sec:metrics}

We evaluate reconstruction fidelity, motion realism, and physical/temporal plausibility.
For distributional realism, we report \textbf{FID} computed on features from an independently trained motion autoencoder~\cite{ling2024versatilemotion}, and \textbf{Diversity (Div)} as the average pairwise distance between randomly sampled generated embeddings.
For reconstruction accuracy, we report \textbf{MPJPE} (mm) and report \textbf{local MPJPE} in parentheses after removing the global root translation.

For kinematic and contact quality, we report \textbf{Skate}~\cite{zhang2024rohm} as the foot skating ratio:
a foot joint is considered near-ground if its height is within $0.08$m above the ground level (estimated as the minimum GT height), and is considered skating if its horizontal speed exceeds $0.1$m/s; Skate is the fraction of near-ground foot samples that satisfy the skating condition.
We also report ground penetration as \textbf{Pen.F} (\%) and \textbf{Pen.D} (mm), computed from toe joints: Pen.F is the percentage of toe samples below the ground by more than $0.05$m, and Pen.D is the mean penetration depth over all toe samples.

For temporal quality, we report \textbf{Acc Err} (m/s$^2$)~\cite{zhang2024rohm}, the mean acceleration difference to GT computed by second-order finite differences scaled by fps, and \textbf{Jitter}~\cite{shen2024world}, computed from third-order finite differences (scaled by fps$^3$ and normalized as in our evaluator).
Finally, we also report \textbf{Keyframe Error (KF Err)}, defined as MPJPE computed \emph{only} on the user-specified control frames (keyframes) used for inbetweening/constraint satisfaction.
\textbf{M2M Score} / \textbf{M2M R@3}~\cite{petrovich2023tmr} measure motion-to-motion retrieval consistency between generated and ground truth motions.

\subsection{Does Multimodal Training Help Even for Noisy-only Denoising?}
\label{sec:ablation_multimodal_denoise}

We further test whether multimodal training improves motion denoising even when \emph{no additional modalities are provided at inference}.
Specifically, we compare C0 (noisy 3D only) and C5 (our full multimodal model) under the same \textbf{auto repair} setting:
both models take the \emph{same} noisy motion as input, and we do not provide interval masks or any extra conditions at test time.
As shown in Table~\ref{tab:multimodal_denoise}, C5 consistently outperforms C0 on reconstruction fidelity and motion quality metrics (e.g., skating, penetration, temporal errors), suggesting that multimodal supervision improves the learned motion prior beyond single-modality training.
We attribute the gains to (i) richer effective training pairings that reduce overfitting to a single modality under limited motion data, (ii) complementary supervisory signals that encourage more accurate conditional modeling during training, and (iii) a more general representation of motion dynamics that transfers back to denoising even when extra cues are absent at inference.

\begin{table}[t]
  \centering
  \caption{\textbf{Multimodal training improves noisy-only denoising.}
  C0 and C5 are evaluated on the same auto repair setup with identical noisy inputs and \emph{no} mask guidance or extra modalities at test time. Lower is better ($\downarrow$) except Div ($\uparrow$).}
  \label{tab:multimodal_denoise}
  \footnotesize
  \setlength{\tabcolsep}{2.8pt}
  \renewcommand{\arraystretch}{1.05}
  \begin{tabular}{@{}l c c c c c c c c@{}}
    \toprule
    Method &
    \shortstack{MPJPE$\downarrow$} &
    \shortstack{Skate$\downarrow$} &
    \shortstack{Acc Err$\downarrow$} &
    \shortstack{Pen.F$\downarrow$} &
    \shortstack{Pen.D$\downarrow$} &
    \shortstack{Jitter$\downarrow$} &
    \shortstack{FID$\downarrow$} &
    \shortstack{Div$\rightarrow$} \\
    \midrule
    GT & 0.00 & 0.13 & 0.00 & 0.00 & 0.00 & 3.73 & 0.00 & 9.40 \\
    \midrule
    \rowcolor{ourblue} C0 & 13.8 & 0.20 & 1.87 & 0.07 & 1.09 & 5.77 & 0.10 & 6.53 \\
    \rowcolor{ourblue} C5 & \textbf{12.8} & \textbf{0.19} & \textbf{1.78} & \textbf{0.03} & \textbf{0.79} & \textbf{5.49} & \textbf{0.06} & \textbf{8.14} \\
    \bottomrule
  \end{tabular}
\end{table}

\subsection{Additional Ablations on Condition Encoders}
\label{sec:ablation_condition_encoders}

We provide additional ablations on how textual and 2D conditions are encoded and injected.
All metrics follow the main paper, and all settings are identical to the corresponding baseline variant unless otherwise specified.

\paragraph{2D condition encoder: lightweight MLP vs.\ MotionBERT features.}
To verify that our model can effectively utilize 2D cues, we ablate the design of the 2D condition encoder while keeping all other components, training settings, and data identical.
Specifically, we compare (i) a lightweight 2D encoder that directly flattens the downsampled 2D keypoints (22$\times$2 per frame) and maps them to the model latent space with a small MLP, versus (ii) a MotionBERT-based 2D feature extractor whose outputs are used as the 2D condition tokens.
As shown in Table~\ref{tab:ablation_2d_encoder}, directly encoding 2D keypoints with a simple MLP consistently outperforms the MotionBERT-based alternative across our evaluation suite.
This indicates that, under our training setup, the model can learn useful 2D-to-motion correspondence from raw keypoints and does not require an explicit 2D-to-3D lifting module before conditioning.
Since the generative backbone already operates in a learned 3D motion latent space, the 2D branch mainly needs to preserve observed image-plane geometry and project it to the model width.
MotionBERT features are stronger for generic 3D pose representation learning, but in our cleanup setting they may introduce a pretrained representation bias and a feature-space mismatch with the autoencoder latents used by the noisy 3D draft and clean 3D keyframes.
The lightweight MLP therefore gives a better trade-off: it preserves direct geometric constraints while letting the cleanup prior infer the compatible 3D motion.

\begin{table}[t]
  \centering
  \caption{\textbf{Ablation on the 2D condition encoder.}
  All settings are identical except for the 2D branch.}
  \label{tab:ablation_2d_encoder}
  \footnotesize
  \setlength{\tabcolsep}{2.8pt}
  \renewcommand{\arraystretch}{1.05}
  \begin{tabular}{@{}l c c c c c c c c@{}}
    \toprule
    Method &
    \shortstack{MPJPE$\downarrow$} &
    \shortstack{Skate$\downarrow$} &
    \shortstack{Acc Err$\downarrow$} &
    \shortstack{Pen.F$\downarrow$} &
    \shortstack{Pen.D$\downarrow$} &
    \shortstack{Jitter$\downarrow$} &
    \shortstack{FID$\downarrow$} &
    \shortstack{Div$\rightarrow$} \\
    \midrule
    GT & 0.00 & 0.13 & 0.00 & 0.00 & 0.00 & 3.73 & 0.00 & 9.40 \\
    \midrule
    \rowcolor{ourblue} C2 (2D MLP; ours) & \textbf{27.45} & \textbf{0.20} & \textbf{2.17} & \textbf{0.07} & \textbf{0.89} & \textbf{6.12} &\textbf{ 0.07} & 8.05 \\
    C2 (MotionBERT 2D)       & 31.45 & 0.21 & 2.24 & 0.10 & \textbf{0.89} & 6.25 & 0.12 & \textbf{9.20} \\
    \bottomrule
  \end{tabular}
\end{table}

\subsection{Ablation on Latent-Space vs.\ Joint-Space 3D Conditioning}
\label{sec:ablation_latent_vs_mlp}

A central design choice in CleanMDM is to encode all 3D geometric conditions (noisy draft and clean keyframes) through the \emph{same} motion autoencoder before fusion, rather than projecting raw joint positions with an MLP.
To isolate the effect of this choice, we compare the full model C5 (CleanMDM) against C5 (MLP), a variant that replaces autoencoder-based 3D keyframe encoding with a lightweight MLP that maps raw joint positions directly to the model width.
All other components --- the flow-matching backbone, text conditioning, 2D keyframe branch --- remain identical.

\begin{table}[t]
  \centering
  \caption{\textbf{Latent-space vs.\ joint-space 3D conditioning (multimodal cleanup).}
  C5 (MLP) replaces the autoencoder-based 3D keyframe encoder with an MLP in joint space.
  All other components are identical to C5.}
  \label{tab:latent_vs_mlp}
  \footnotesize
  \setlength{\tabcolsep}{2.4pt}
  \renewcommand{\arraystretch}{1.05}
  \begin{tabular}{@{}l c c c c c c c c c@{}}
    \toprule
    Method &
    \shortstack{MPJPE$\downarrow$} &
    \shortstack{Skate$\downarrow$} &
    \shortstack{Acc Err$\downarrow$} &
    \shortstack{Pen.F$\downarrow$} &
    \shortstack{Pen.D$\downarrow$} &
    \shortstack{Jitter$\downarrow$} &
    \shortstack{KF Err$\downarrow$} &
    \shortstack{FID$\downarrow$} &
    \shortstack{Div$\rightarrow$} \\
    \midrule
    GT & 0.00 & 0.13 & 0.00 & 0.00 & 0.00 & 3.73 & 0.00 & 0.00 & 9.40 \\
    \midrule
    C5 (MLP)
      & 35.59 & \textbf{0.10} & 14.94 & 0.08 & 0.73 & 71.02 & 30.44 & 2.99 & 8.91 \\
    \rowcolor{ourblue}
    C5 (CleanMDM)
      & \textbf{21.95} & 0.18 & \textbf{1.98} & \textbf{0.06} & \textbf{0.76} & \textbf{5.54} & \textbf{11.4} & \textbf{0.11} & \textbf{9.23} \\
    \bottomrule
  \end{tabular}
\end{table}

As shown in Table~\ref{tab:latent_vs_mlp}, C5 (MLP) degrades dramatically compared to C5 on temporal quality (Jitter: 71.02 vs.\ 5.54, Acc: 14.94 vs.\ 1.98) and distributional realism (FID: 2.99 vs.\ 0.11), while also showing worse reconstruction accuracy (MPJPE: 35.59 vs.\ 21.95) and keyframe adherence (KF Err: 30.44 vs.\ 11.4).
Three observations emerge:

\paragraph{(i) The AE bottleneck provides a temporal-coherence prior.}
The autoencoder's temporal downsampling ($d{=}4$) forces the encoder to produce representations that are inherently smooth across frames.
An MLP encoding each frame independently lacks this inductive bias.
As a result, C5 (MLP) can match individual keyframe positions but produces incoherent transitions between them, explaining the catastrophic jitter ($12.8\times$) and acceleration error ($7.5\times$).

\paragraph{(ii) Shared representation space stabilizes weighted fusion.}
In C5, both the noisy 3D draft and clean 3D keyframes are encoded by the same autoencoder, so the confidence-weighted sum in Eq.~(2) operates in a common latent space with aligned scale and temporal rate.
In C5 (MLP), the MLP-encoded keyframe tokens occupy a different representation space from the AE-encoded noisy draft, creating a modality gap that destabilizes fusion and undermines conditioning quality.

\paragraph{(iii) Distributional quality requires representation-level alignment.}
The $27\times$ FID degradation indicates that raw-space encoding does not merely reduce point-wise accuracy --- it fundamentally disrupts the generated motion distribution.
C5 (MLP) produces motions that deviate sharply from natural motion statistics, even though its skating ratio is nominally lower (0.10 vs.\ 0.18), suggesting that the low skating comes at the cost of unnatural, jerky motion rather than genuine physical plausibility.

\begin{table}[t]
  \centering
  \caption{\textbf{Ablation on LMQD feature representations (denoising).}
  All variants use the same denoising setup; only the discriminator feature is changed.}
  \label{tab:ablation_lmqd_features}
  \footnotesize
  \setlength{\tabcolsep}{3.2pt}
  \renewcommand{\arraystretch}{1.05}
  \begin{tabular}{@{}l c c c c c c@{}}
    \toprule
    Method &
    \shortstack{MPJPE$\downarrow$} &
    \shortstack{Skate$\downarrow$} &
    \shortstack{Acc Err$\downarrow$} &
    \shortstack{Pen.F$\downarrow$} &
    \shortstack{Pen.D$\downarrow$} &
    \shortstack{Jitter$\downarrow$} \\
    \midrule
    \rowcolor{ourblue} C0 (diffusion only)                  & 13.8 & 0.2008 & \textbf{1.87} & 0.07 & 1.09 & 6.5286 \\
    \rowcolor{ourblue} C0+LMQD (latent seq, ours)           & \textbf{11.8} & \textbf{0.1780} & 1.96 & \textbf{0.02} & \textbf{0.46} & \textbf{6.3115} \\
    C0+LMQD (power seq)                  & 13.3 & 0.1813 & 1.96 & 0.05 & 0.91 & 6.4056 \\
    C0+LMQD (joint seq)                  & 13.3 & 0.1843 & 2.01 & 0.04 & 0.65 & 6.5149 \\
    C0+LMQD (AMP seq)                    & 13.3 & 0.1847 & 2.01 & 0.04 & 0.69 & 6.5257 \\
    C0+LMQD (accel seq)                  & 13.3 & 0.1848 & 2.01 & 0.04 & 0.66 & 6.5167 \\
    \bottomrule
  \end{tabular}
\end{table}

\begin{table}[!t]
\centering
\caption{
\textbf{Generalization on public datasets.}
We report MPJPE to the real-world input motion (lower indicates less drift), together with temporal smoothness (Jitter) and contact quality (Slide R/A).
These public datasets do not provide ground-truth cleaned motions.
CleanMDM consistently reduces sliding and jitter while maintaining fidelity to the real-world input motion, indicating a transferable motion prior that improves cleanup without over-smoothing.}
\footnotesize
\setlength{\tabcolsep}{3.2pt}
\renewcommand{\arraystretch}{1.05}

\begin{tabular}{@{}l l c c c c@{}}
\toprule
Dataset & Method & MPJPE (mm)$\downarrow$ & Jitter$\downarrow$ & Slide R$\downarrow$ & Slide A$\downarrow$ \\
\midrule \midrule
\multirow{3}{*}{IDEA400}
& Input & 0.00 & 5.43 & 0.30 & 0.06 \\
& StableMotion & 40.10 & 5.10 & 0.45 & 0.14 \\
& \cellcolor{ourblue}\textbf{CleanMDM} 
    & \cellcolor{ourblue}\textbf{15.61} 
    & \cellcolor{ourblue}\textbf{4.94} 
    & \cellcolor{ourblue}\textbf{0.25} 
    & \cellcolor{ourblue}\textbf{0.05} \\
\midrule
\multirow{3}{*}{Animation}
& Input & 0.00 & 17.60 & 0.54 & 0.15 \\
& StableMotion & 138.61 & \textbf{11.27} & 0.60 & 0.11 \\
& \cellcolor{ourblue}\textbf{CleanMDM} 
    & \cellcolor{ourblue}\textbf{69.40} 
    & \cellcolor{ourblue}15.95 
    & \cellcolor{ourblue}\textbf{0.50} 
    & \cellcolor{ourblue}\textbf{0.11} \\
\midrule
\multirow{3}{*}{HAA500}
& Input & 0.00 & 20.78 & 0.62 & 0.14 \\
& StableMotion & 147.50 & \textbf{10.93} & 0.58 & 0.11 \\
& \cellcolor{ourblue}\textbf{CleanMDM} 
    & \cellcolor{ourblue}\textbf{57.83} 
    & \cellcolor{ourblue}12.87 
    & \cellcolor{ourblue}\textbf{0.53}
    & \cellcolor{ourblue}\textbf{0.10} \\
\midrule
\multirow{3}{*}{HuMMan}
& Input & 0.00 & 10.30 & 0.41 & 0.08 \\
& StableMotion & \textbf{84.02} & 10.17 & 0.36 & 0.07 \\
& \cellcolor{ourblue}\textbf{CleanMDM}
  & \cellcolor{ourblue}84.72
  & \cellcolor{ourblue}\textbf{7.80}
  & \cellcolor{ourblue}\textbf{0.32}
  & \cellcolor{ourblue}\textbf{0.06} \\
\midrule
\multirow{3}{*}{KungFu}
& Input & 0.00 & 16.63 & 0.62 & 0.12 \\
& StableMotion & 127.84 & \textbf{9.15} & 0.56 & 0.11 \\
& \cellcolor{ourblue}\textbf{CleanMDM} 
    & \cellcolor{ourblue}\textbf{69.40} 
    & \cellcolor{ourblue}11.95 
    & \cellcolor{ourblue}\textbf{0.56} 
    & \cellcolor{ourblue}\textbf{0.11} \\
\bottomrule
\end{tabular}
\label{tab:public_datasets}
\end{table}

\subsection{Ablation on LMQD Feature Representations}
\label{sec:ablation_lmqd_features}

In the main paper, we introduce the \emph{Latent Motion Quality Discriminator} (\textbf{LMQD}) as a lightweight adversarial regularizer that operates on entire sequences and improves kinematic realism beyond regression-style objectives.
Here we ablate the \emph{feature representation} fed to LMQD to verify that the latent sequence is the most effective choice.

\paragraph{Setup.}
We isolate this factor in a pure denoising setting (blind auto repair): all variants share the same diffusion backbone, training protocol, and noisy inputs, and differ only in the discriminator feature.
We compare a diffusion-only baseline (\textbf{C0}) against \textbf{C0+LMQD} with different discriminator inputs:
(i) \textbf{latent seq} (ours), and several alternatives computed from joint-space motion statistics, including
(ii) \textbf{power seq}, (iii) \textbf{joint seq}, (iv) \textbf{AMP seq}, a handcrafted state feature sequence inspired by the adversarial motion prior used in PHC/AMP-style pipelines~\cite{luo2023perpetual,peng2021amp} (e.g., concatenating joint positions/velocities and contact-related signals over time), and (v) \textbf{accel seq}

\paragraph{Results.}
Table~\ref{tab:ablation_lmqd_features} shows that using the \emph{latent sequence} as the LMQD input provides the best overall trade-off.
Compared to \textbf{C0}, \textbf{C0+LMQD (latent seq)} consistently reduces foot skating and ground penetration while improving MPJPE, and it outperforms alternative feature choices under the same training budget.
This supports the design used throughout our method: the autoencoder latent offers a compact, temporally aligned representation that is well-suited for sequence-level discrimination, yielding a stronger distributional constraint on motion quality.

\subsection{Generalization to Public Datasets}
\label{sec:public}

\paragraph{Cleanup on out-of-domain motions.}
We evaluate on five public datasets spanning different capture modalities.
\textbf{IDEA400}~\cite{lin2023motion} and \textbf{KungFu}~\cite{lin2023motion} are monocular video-based motion capture subsets from Motion-X, covering everyday actions and martial arts sequences respectively.
\textbf{HAA500}~\cite{chung2021haa500} is a monocular video dataset containing 500 human action categories.
\textbf{HuMMan}~\cite{cai2022humman} is a multi-view RGB-D dataset with higher 3D reconstruction accuracy due to synchronized multi-camera capture.
\textbf{Animation}~\cite{ling2024versatilemotion} consists of artist-authored keyframe animation data; unlike video-based sources, it exhibits inherently higher jitter values owing to the discrete keyframe interpolation process rather than capture noise.
Table~\ref{tab:public_datasets} evaluates our best cleanup model (without Mesh-Aware Contact Projection) on several public datasets and compares it to StableMotion.
Across datasets, our method consistently reduces sliding and jitter relative to StableMotion, while keeping the motion closer to the real-world input motion (MPJPE).
This suggests that CleanMDM learns a transferable motion prior that improves temporal stability and contact behavior without over-smoothing or drifting the underlying performance.

\begin{figure*}[t]
    \centering
    \includegraphics[width=0.95\linewidth]{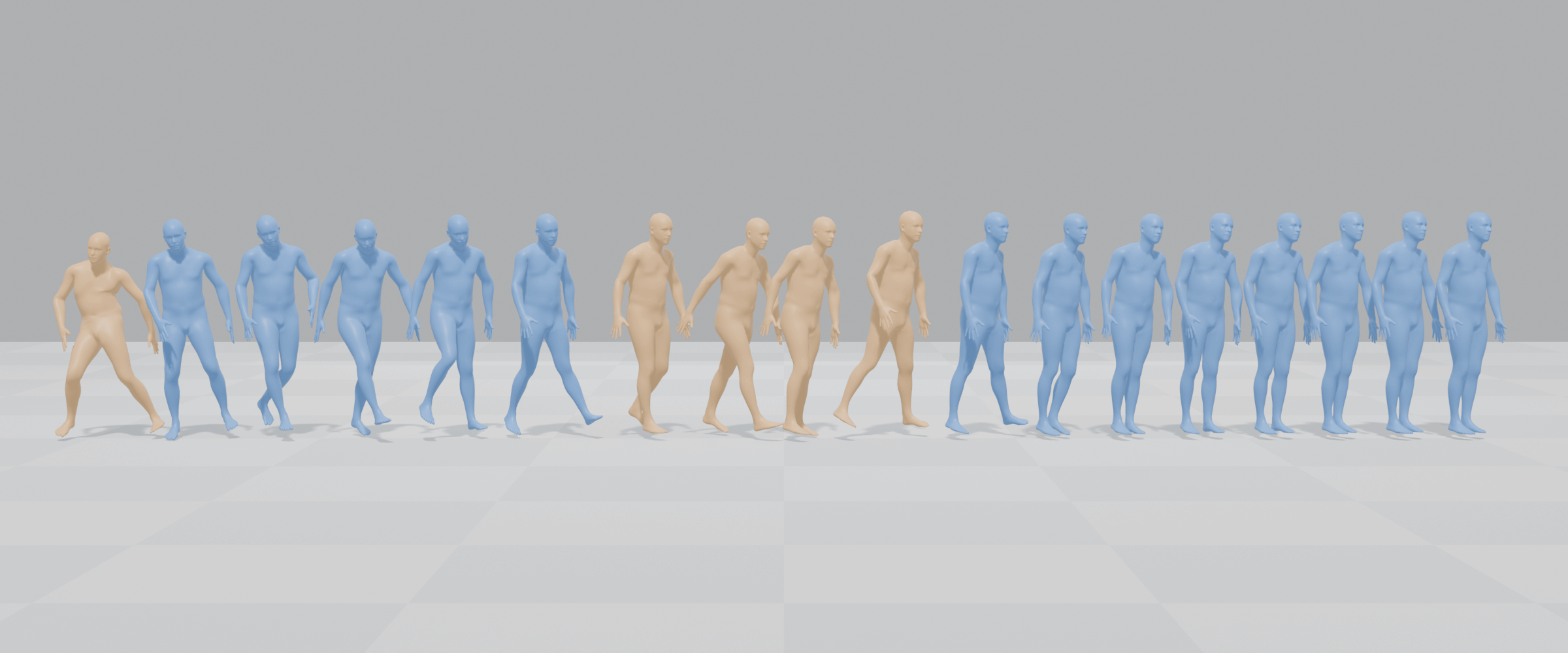}\\
    \includegraphics[width=0.95\linewidth]{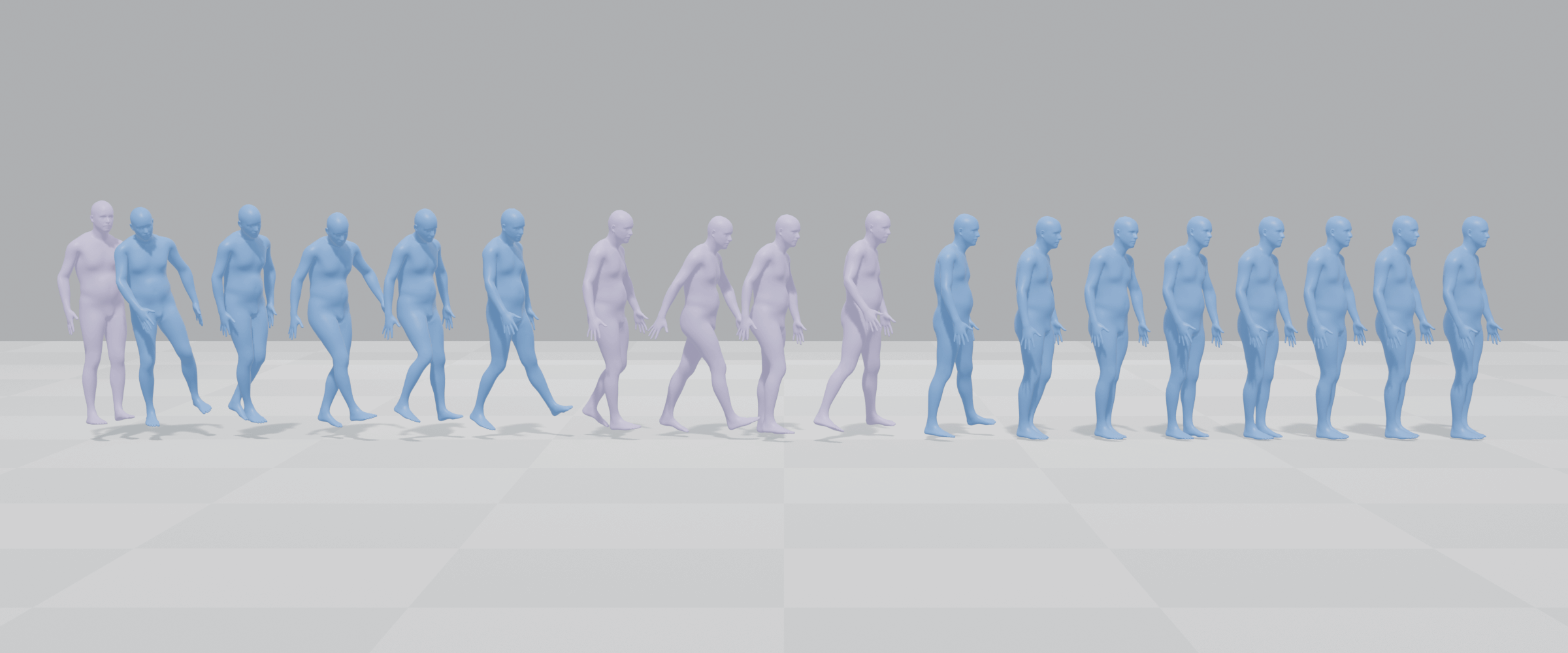}\\
    \includegraphics[width=0.95\linewidth]{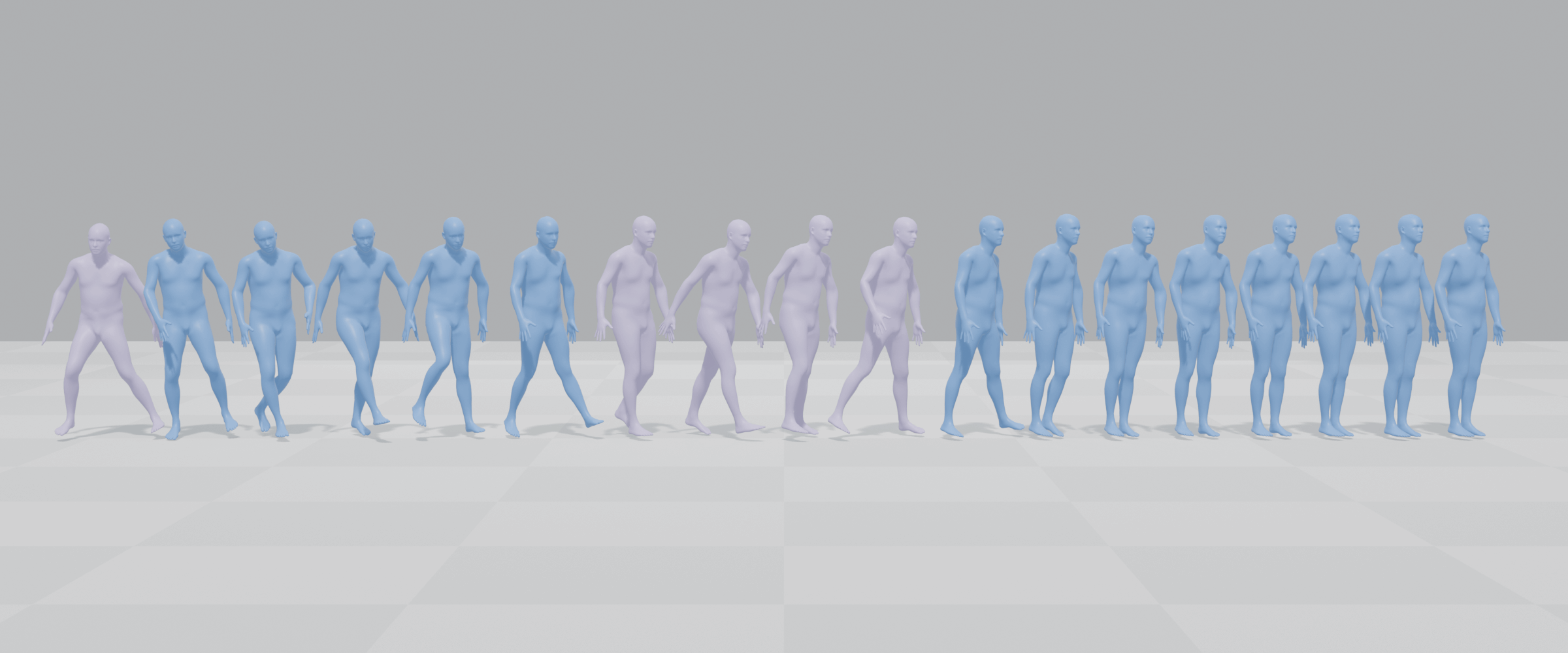}
    \caption{\textbf{(a) Real-world cleanup on KungFu \emph{Tornado Kick} (Motion-X)~\cite{lin2023motion}.} Top to bottom: real-world input, StableMotion~\cite{mu2025stablemotion}, and \textbf{CleanMDM (ours)}. Orange bodies mark low-quality frames in the real-world input, while purple bodies mark the corresponding repaired frames in StableMotion and CleanMDM. StableMotion shows visible support-foot floating and dampened rotation; CleanMDM closely tracks the input motion. See the preceding paragraph for the MPJPE/Jitter trade-off, and Fig.~\ref{fig:kungfu_basketball} for a second example on the same dataset.}
    \label{fig:kungfu_unfolded}
\end{figure*}

\begin{figure*}[t]
    \centering
    \includegraphics[width=0.95\linewidth]{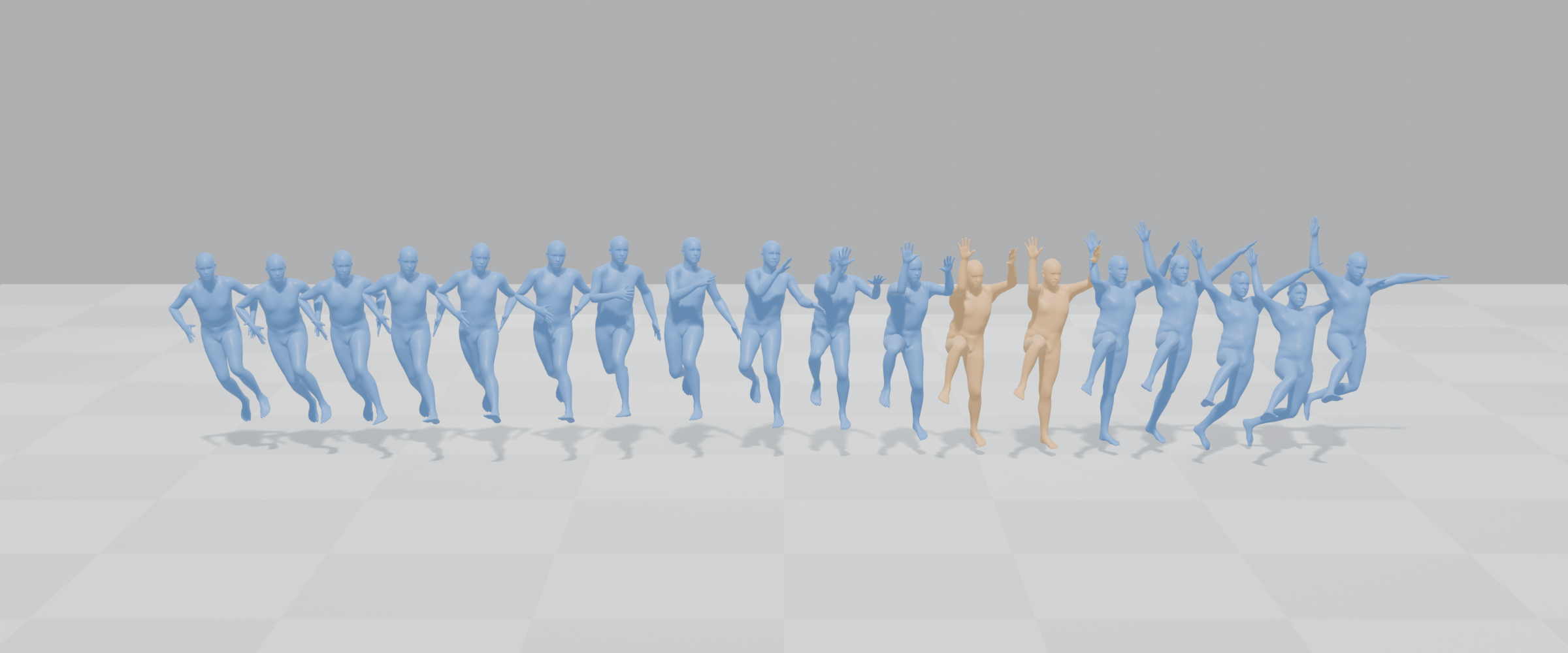}\\
    \includegraphics[width=0.95\linewidth]{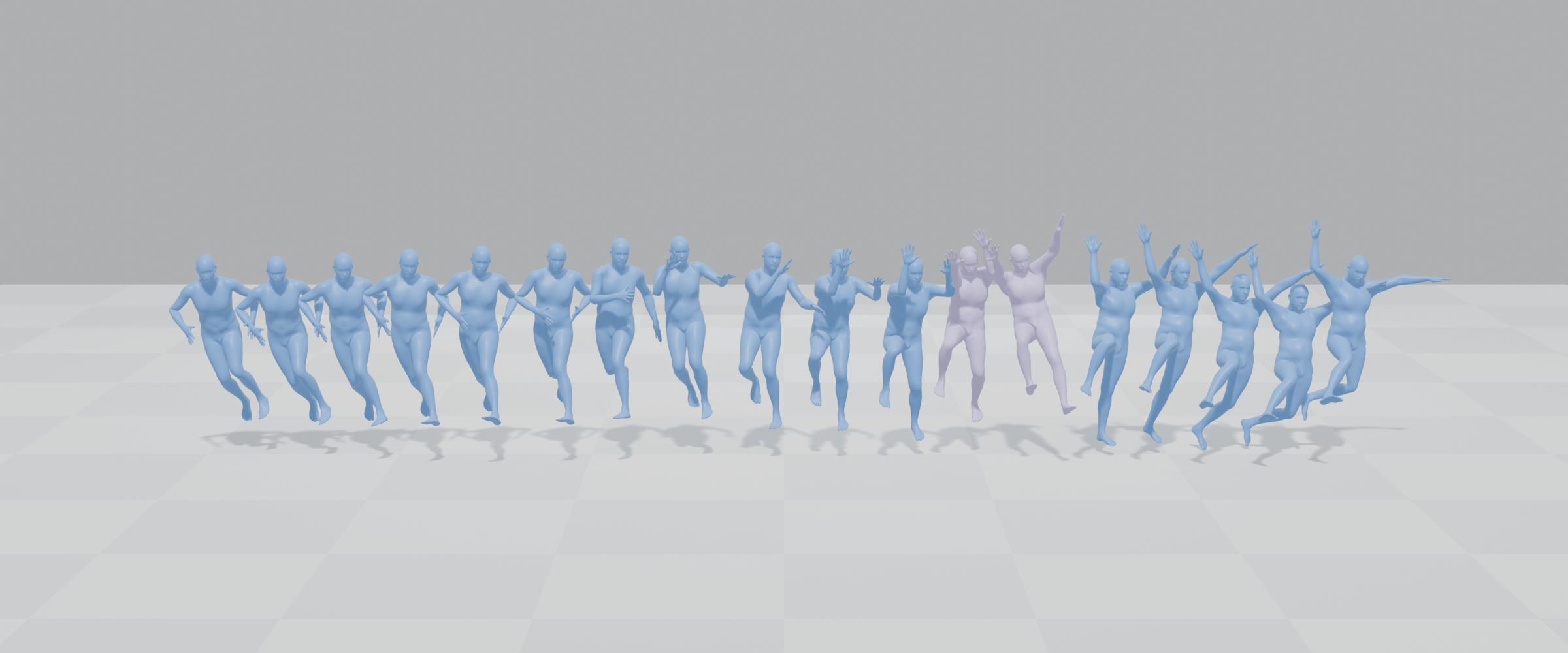}\\
    \includegraphics[width=0.95\linewidth]{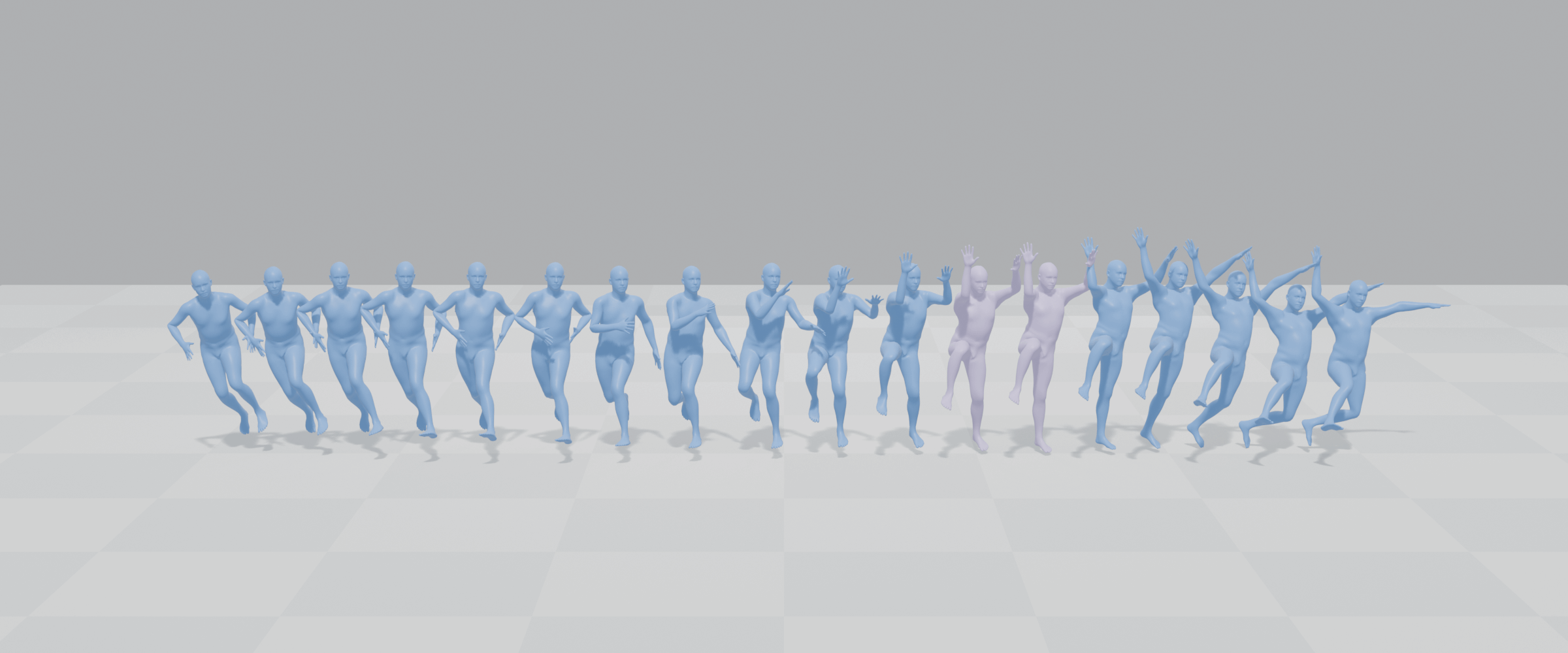}
    \caption{\textbf{(b) Additional KungFu example: \emph{Play basketball}.} Top to bottom: real-world input, StableMotion~\cite{mu2025stablemotion}, and \textbf{CleanMDM (ours)}. Orange bodies mark low-quality frames in the real-world input, while purple bodies mark the corresponding repaired frames in StableMotion and CleanMDM. StableMotion lifts the subject off the ground in one of these frames, breaking the grounded action semantics; CleanMDM preserves the input contact behavior. Companion to Fig.~\ref{fig:kungfu_unfolded}.}
    \label{fig:kungfu_basketball}
\end{figure*}

\paragraph{KungFu visualization and the Jitter trade-off.}
Figures~\ref{fig:kungfu_unfolded} and~\ref{fig:kungfu_basketball} present two representative KungFu (Motion-X) clips, (a) \emph{Tornado Kick} and (b) \emph{Play basketball}, comparing the real-world input, StableMotion, and CleanMDM as horizontally-unfolded SMPL mesh sequences.
The figures make the Jitter row of Table~\ref{tab:public_datasets} interpretable: StableMotion's lower Jitter on KungFu (9.2 vs.\ 12.0) is achieved by aggressively smoothing the temporal trajectory, which simultaneously detaches the support foot from the ground (floating) and weakens the kick's rotational signature, yielding a much higher MPJPE (127.8 vs.\ 69.4).
CleanMDM instead preserves the action's spatial structure and contact behavior, at the cost of a slightly higher numerical jitter that mirrors the high-frequency content of the input motion itself.
This is consistent with the role of CleanMDM as a generative cleanup prior conditioned on the input rather than a temporal smoother.

Since jitter is difficult to assess from static unfolded meshes, Figs.~\ref{fig:kungfu_unfolded} and~\ref{fig:kungfu_basketball} focus on visible grounding/contact artifacts such as floating and foot-ground consistency. Together, these static examples show that CleanMDM preserves the action semantics while reducing penetration/floating and producing smoother motion with less visible jitter.

\subsection{User Study}
\label{sec:userstudy_appendix}

\paragraph{Protocol.}
We conduct a perceptual study to evaluate whether CleanMDM improves motion quality in ways that are salient to animators.
We consider three practical tasks: \textbf{auto cleanup}, \textbf{inbetween}, and \textbf{multimodal cleanup}.
For each task, we randomly sample $10$ clips from the test set and compare CleanMDM against a task-matched baseline
(StableMotion for auto cleanup, CondMDI for inbetween, and GenMo for multimodal cleanup).
For each clip, we present a \emph{paired} comparison (CleanMDM vs.\ baseline) with randomized left/right order and anonymized method names.

\paragraph{Questions.}
To reduce rater fatigue while capturing production-relevant qualities, raters score each method on three criteria:
\textbf{Artifacts}, \textbf{Naturalness}, and \textbf{Faithfulness}.
Artifacts summarize contact and temporal issues (e.g., foot skating, ground penetration, and jitter).
Naturalness evaluates whether the motion looks physically and stylistically plausible.
Faithfulness measures consistency with the reference (ground truth for our benchmark; or the provided constraints for multimodal cleanup).
All scores are given on a $0$--$5$ scale, where higher is better.

\paragraph{Participants and results.}
We recruit $10$ professional animators to perform the study.
All participants are colleagues who volunteered; no monetary compensation was provided.
Table~\ref{tab:user_study} reports mean ratings aggregated over all clips and participants for each task.
CleanMDM is consistently preferred over task-matched baselines across all three criteria, with the largest gains observed in \emph{Faithfulness} and \emph{Artifacts}.

\begin{table*}[t]
\centering
\caption{\textbf{Complete user-study results.} Mean ratings from $10$ professional animators. Higher is better.}
\label{tab:user_study}
\footnotesize
\setlength{\tabcolsep}{3.0pt}
\renewcommand{\arraystretch}{1.08}
\begin{tabular}{@{}l l c c c@{}}
  \toprule
  Setting & Method &
  Artifacts$\uparrow$ & Naturalness$\uparrow$ & Faithfulness$\uparrow$ \\
  \midrule \midrule
  \multirow{2}{*}{Auto cleanup}
  & StableMotion
    & 3.45 & 3.61 & 2.33 \\
  & \cellcolor{ourblue}\textbf{CleanMDM}
    & \cellcolor{ourblue}\textbf{4.13}
    & \cellcolor{ourblue}\textbf{4.41}
    & \cellcolor{ourblue}\textbf{4.63} \\
  \midrule
  \multirow{2}{*}{Inbetween}
  & CondMDI
    & 2.91 & 3.08 & 2.29 \\
  & \cellcolor{ourblue}\textbf{CleanMDM}
    & \cellcolor{ourblue}\textbf{3.93}
    & \cellcolor{ourblue}\textbf{4.07}
    & \cellcolor{ourblue}\textbf{4.33} \\
  \midrule
  \multirow{2}{*}{Multimodal cleanup}
  & GenMo
    & 3.20 & 3.50 & 3.66 \\
  & \cellcolor{ourblue}\textbf{CleanMDM}
    & \cellcolor{ourblue}\textbf{3.93}
    & \cellcolor{ourblue}\textbf{4.12}
    & \cellcolor{ourblue}\textbf{4.22} \\
  \bottomrule
\end{tabular}
\end{table*}


\end{document}